\documentclass[conference]{IEEEtran}
\IEEEoverridecommandlockouts
\usepackage{amssymb, mathtools, xargs, units, cite, mathrsfs, graphicx, xcolor, amsmath, listings, tikz, pgfplots, cancel, multicol, geometry, multirow, booktabs,amsthm, pgfplotstable}
\pgfplotsset{compat=1.18}
\usepgfplotslibrary{groupplots,fillbetween,colormaps}
\pgfplotsset{compat=newest}

\definecolor{customgreen}{rgb}{0.0, 0.6, 0.0}

\usetikzlibrary{3d, calc}
\usetikzlibrary{arrows.meta}

\newcommand{\clip}{\operatorname{clip}}

\usepackage{amsmath,amssymb,mathtools}
\usepackage{booktabs}
\usepackage{multirow}
\usepackage{graphicx}
\usepackage{xcolor}
\usepackage{url}
\usepackage{cite}
\usepackage{array}
\usepackage{tabularx}
\usepackage{balance}
\usepackage{tikz}
\usetikzlibrary{arrows.meta,positioning,fit}

\allowdisplaybreaks
\title{\LARGE Exposure-Aware Informative Path Planning Using Deep Reinforcement Learning}
\author{Muzaffar Qureshi$^{1}$ \and Yihui Mao$^{1}$  \and James W. Cross$^{1}$ \and Kyle Volle$^{2}$ \and Rushikesh Kamalapurkar$^{1}$%
\thanks{This research was supported in part by the Air Force Research Laboratory under contract number FA8651-24-1-0019. Any opinions, findings, or recommendations in this article are those of the author(s), and do not reflect the views of the sponsoring agencies.}%
\thanks{$^{1}$ Department of Mechanical and Aerospace Engineering, University of Florida, Gainesville, FL, USA, email: {\tt\footnotesize \{muzaffar.qureshi, yihui.mao,jamescross, rkamalapurkar\} @ufl.edu}.}%
\thanks{$^{2}$ Torch Technologies, Shalimar, FL, USA, email: {
\tt \footnotesize Kyle.Volle@torchtechnologies.com}}}

\begin{document}
\maketitle
\pagestyle{empty}

\begin{abstract}
This paper addresses the finite-budget exposure-aware informative path planning (IPP) problem for a fleet of mobile sensing agents. The agents map an unknown scalar field while limiting exposure to high field intensity. Existing IPP methods optimize informativeness under motion or resource constraints, but do not address field exposure constraints. We formulate exposure-aware IPP (EA-IPP) as a scalarized bi-objective belief-state Markov decision process that trades the normalized Gaussian-process (GP) information gain against a normalized measurement-derived fleet exposure index. A parameter-shared proximal policy optimization (PPO) framework with decentralized actors and a centralized fleet critic is trained offline and evaluated on unseen fields. Across the radar, plume, and wildfire field families, the EA-IPP agent achieves the lowest GP reconstruction RMSE among the evaluated baselines while retaining D-optimal information within $10\%$ of the non-learning information baseline. The minimum field exposure is approximately $34.6\%$, $20.1\%$, and $27\%$ below the non-exposure-aware baseline for radar, plume, and wildfire, respectively.
\end{abstract}

\section{Introduction}
\label{sec:introduction}
Autonomous robots are useful tools for mapping spatial phenomena such as pollutant concentrations, radiological dose rates, water-quality indicators, thermal fields, biological abundance, and electromagnetic fields~\cite{SCC.Singh.Krause.ea2007,SCC.Popovic.VidalCalleja.ea2020}. Robot trajectories must be planned to visit informative measurement locations while respecting the operational limitations of the robot~\cite{SCC.Hollinger.Sukhatme2014}. In applications where exposure to the field is detrimental, the fleet must select a \emph{sequence} of measurement locations that improves field map accuracy while limiting field exposure under finite energy or time budgets~\cite{SCC.Singh.Krause.ea2007,SCC.Arora.Scherer2017}.

Information maximization under motion and resource constraints is studied under the moniker of informative path planning (IPP)~\cite{SCC.Binney.Sukhatme2012,SCC.Hollinger.Sukhatme2014}. Existing IPP methods include combinatorial and sampling-based planning, trajectory and Bayesian optimization, and receding-horizon planning~\cite{SCC.Hollinger.Sukhatme2014,SCC.Arora.Scherer2017, hitz2017adaptive,marchant2014bayesian}. In adaptive IPP, each measurement updates the field belief, changing the estimated informativeness of future measurement locations. These estimates are then incorporated into sampling-, optimization-, or learning-based planners through online or receding-horizon replanning~\cite{hitz2017adaptive,popovic2024survey}. A probabilistic belief, such as a Gaussian process (GP), is therefore central to adaptive IPP~\cite{rasmussen2006gaussian}. Bayesian active design can select measurements from predictive uncertainty~\cite{muzaffar.MacKay1992}, while GP experimental design uses A- and D-optimality to reduce posterior uncertainty through trace- and determinant-based criteria, respectively~\cite{chaloner1995bayesian,krause2008near}. 

Covariance-based information criteria depend on measurement locations, not field measurements. Consequently, two measurement locations can provide comparable uncertainty reduction while inducing different exposure. The information-exposure trade-off considered in this paper requires an objective that jointly accounts for posterior uncertainty reduction and the exposed field intensity. The GP posterior covariance identifies locations where additional measurements can reduce model uncertainty, whereas the posterior field prediction identifies which future measurements are likely to induce high field exposure.

In cases where the GP is updated after each measurement, and each action affects future fleet configurations and trajectories, EA-IPP is a finite-horizon sequential decision problem. A reinforcement learning (RL) feedback policy is therefore trained offline to map the current belief and fleet state to the next action, avoiding long-horizon action-sequence search at deployment~\cite{popovic2024survey}.

Yanes Luis \emph{et al.} considered multi-agent ASV patrolling using deep Q-learning~\cite{yanes2021multiagent}, and later incorporated local GP mean and uncertainty estimates into multi-agent DRL for information gathering~\cite{SCC.YanesLuis.Shutin.ea2024}. These approaches optimize coverage or information gain, but do not account for cumulative exposure to the measured field. Related safe-RL methods address risk through constrained objectives or safety mechanisms~\cite{garcia2015survey,achiam2017cpo}, but are not formulated around the joint information--exposure objective considered here.

Constrained Bayesian optimization methods restrict motion actions to regions satisfying prescribed safety constraints~\cite{sui2015safe,turchetta2016safe}, while related safe GP approaches estimate unsafe sets and redirect candidate locations to the estimated safe region~\cite{Qureshi2024ScalarFieldMapping,Qureshi2026HoughTransform}. These methods require a prescribed safety threshold to define the feasible set, whereas exposure-aware planning can instead trade information value against increasing field intensity without imposing a cutoff threshold.

In this work, exposure is traded against information acquisition through a hedging parameter $\lambda\in[0,1]\subset\mathbb{R}$, so high-intensity regions remain admissible to gain information value while incurring an exposure penalty. Finite-budget multi-agent IPP is formulated as a multi-objective belief-state Markov decision process (MDP), and an RL policy is trained offline on simulated fields for evaluation on unseen fields without requiring long-horizon combinatorial search. The contributions are as follows.
\begin{enumerate}
    \item The formulation of an exposure-aware belief-state MDP for finite-budget IPP that couples covariance-based information acquisition with field exposure.

    \item A reward design in which $\lambda$ provides a mission-level trade-off between information gain and exposure.
    
    \item The information-seeking PPO policy achieves lower RMSE than the evaluated information-driven baselines, while exposure weighting reduces exposure relative to information-seeking PPO.
\end{enumerate}

\section{Problem Formulation}
\label{sec:problem}
Let $\mathcal{X}\subset\mathbb{R}^{2}$ denote the domain of interest, and $f:\mathcal{X}\rightarrow\mathbb{R}_{\geq 0}$ denote an unknown scalar field. The objective is to estimate $f$ from noisy measurements over a finite horizon while limiting field exposure. The domain is discretized into a uniform grid with axis-aligned spacing $\Delta>0$ along each coordinate direction. The subset reachable by an agent, excluding boundary and obstacle cells, is the navigable grid $\mathcal{X}_{\mathrm g}=\{\mathbf{x}^{(1)},\ldots,\mathbf{x}^{(n_c)}\}\subset\mathcal X$, with $|\mathcal X_{\mathrm g}|=n_c$. At each decision step, agent $i$ selects a feasible action and moves to a neighboring grid point in $\mathcal{X}_{\mathrm g}$, where the next measurement is acquired.

Let $\mathcal{V}:=\{1,\ldots,N\}$ denote a fleet of $N\in\mathbb N$ mobile sensing robots. The mapping episode has a fixed horizon of $H\in\mathbb N$ steps. The position and measurement indices of the agents satisfy $t\in\{0,\ldots,H\}$, whereas action and reward indices satisfy $t\in\{0,\ldots,H-1\}$, where $t=0$ denotes the initial locations and measurements of the robots.

The position of robot $i$ is denoted by $\mathbf {p}_{i,t}\in\mathcal X_{\mathrm g}$, and $\mathbf X_t:=(\mathbf p_{1,t},\ldots,\mathbf p_{N,t})\in\mathcal X_{\mathrm g}^{N}$ denotes the fleet positions at step $t$. Let $\mathcal A:=\{1,\ldots,9\}$ denote the finite action set. Each action, denoted as $a\in\mathcal A$, is associated with a local displacement through the bijection $\mathbf d:\mathcal A\rightarrow\{-1,0,1\}^{2}$, where $\mathbf d(a)=(d_x(a),d_y(a))$. At step $t$, robot $i$ selects an action $a_{i,t}\in\mathcal A$. Here, $d_x(a),d_y(a)\in\{-1,0,1\}$ denote the grid displacements along the $x$- and $y$-directions, respectively. For a given $t$, the actions selected by all agents form the joint action vector
$\mathbf a_t := (a_{1,t},\ldots,a_{N,t}) \in\mathcal A^N$. Executing action $a_{i,t}$ updates the position according to
\begin{equation}
\mathbf p_{i,t+1}
=
\mathbf p_{i,t}
+
\Delta\,\mathbf d(a_{i,t}).
\label{eq:state_transition}
\end{equation}
The subset of diagonal actions is defined by $\mathcal A_{\mathrm{diag}} := \{a\in\mathcal A : d_x(a)d_y(a)\neq 0\}$. The set of navigation-feasible actions for agent $i$ at step $t$, excluding map boundaries, is defined as \begin{equation}\scalebox{0.9}{$
\mathcal A_{\mathrm{nav},i,t}
:=
\left\{
a\in\mathcal A :
\begin{aligned}
&\mathbf p_{i,t}
+\Delta\,\mathbf d(a)
\in \mathcal X_{\mathrm g}, \\
&a\in\mathcal A_{\mathrm{diag}}
\Rightarrow
\mathbf p_{i,t}
+\Delta\bigl(d_x(a),0\bigr)
\in \mathcal X_{\mathrm g}, \\
&a\in\mathcal A_{\mathrm{diag}}
\Rightarrow
\mathbf p_{i,t}
+\Delta\bigl(0,d_y(a)\bigr)
\in \mathcal X_{\mathrm g}
\end{aligned}
\right\}.$}
\label{eq:navigation_action_set}
\end{equation}
Let $\mathbf a=(a_1,\ldots,a_N)\in\prod_{i\in\mathcal V}\mathcal A_{\mathrm{nav},i,t}$ denote a joint action, where $\prod_{i\in\mathcal V}\mathcal A_{\mathrm{nav},i,t}$ is the set of all joint actions satisfying the individual navigation constraints. The admissible joint action set $\mathcal A_{\mathrm{feas}}(\mathbf X_t)$ consists of all $\mathbf a$ such that, for every $i\neq j$, $\mathbf q_i\neq\mathbf q_j$ (no vertex conflict), $(\mathbf q_i,\mathbf q_j)\neq(\mathbf p_{j,t},\mathbf p_{i,t})$, and the interiors of the motion segments $[\mathbf p_{i,t},\mathbf q_i]$ and $[\mathbf p_{j,t},\mathbf q_j]$ do not intersect. The executed joint action satisfies $\mathbf a_t\in\mathcal A_{\mathrm{feas}}(\mathbf X_t)$. Revisiting previously measured locations is permitted, and the wait action remains available whenever no motion action is feasible.

At each measurement instance $t$, agent $i$ acquires a noisy measurement according to
\begin{equation}
y_{i,t}
=
f(\mathbf p_{i,t})
+
\varepsilon_{i,t},
\label{eq:obs_model}
\end{equation} where $\varepsilon_{i,t} \overset{\mathrm{i.i.d.}}{\sim} \mathcal {N}(0,\sigma_n^2)$ and $\sigma_n^2>0$ is the sensor noise variance. The measurements collected by all agents form the multiset $\mathcal B_t:=\biguplus_{i\in\mathcal V}\{(\mathbf p_{i,t},y_{i,t})\}$, with $|\mathcal B_t|=N$ \footnote{For multisets $A_i$, $\biguplus_{i\in\mathcal V} A_i$ contains each element $x$ with multiplicity $\sum_{i\in\mathcal V} m_{A_i}(x)$, where $m_{A_i}(x)$ denotes the multiplicity of $x$ in $A_i$.}. For $t\in\{0,\ldots,H\}$, the pooled measurement history is $\mathcal D_t := \biguplus_{\tau=0}^{t}\mathcal B_\tau$, where $\mathcal B_0$ denotes the initial positions and measurements of the fleet.

Let $\Sigma_t\in\mathbb S_{++}^{n_c}$ denote the global GP posterior covariance over $\mathcal X_{\mathrm g}$ after assimilating $\mathcal D_t$, derived in Section~\ref{sec:method}. The information gain associated with joint action $\mathbf a_t$ and the subsequent measurement batch $\mathcal B_{t+1}$ is denoted as
\begin{equation}
I_t := \mathcal U(\Sigma_t) - \mathcal U(\Sigma_{t+1}),
\label{eq:info_gain}
\end{equation}
where $\mathcal U:\mathbb S_{++}^{n_c}\to\mathbb R$ is a posterior-uncertainty function instantiated in Section~\ref{sec:reward} through A- or D-optimality. 

Let $\mathcal M_t := (\mathbf X_t,\mathcal D_t)$ denote the information state at step $t$. The initial fleet configuration $\mathbf X_0$ is drawn from a distribution $\mathbb P_0$ over collision-free configurations in $\mathcal X_{\mathrm g}^{N}$. At each step, the robots select a joint action $\mathbf a_t\in\mathcal A_{\mathrm{feas}}(\mathbf X_t)$, where
$\mathcal A_{\mathrm{feas}}(\mathbf X_t)$ contains the joint actions that satisfy the individual navigation constraints and the motion constraints; joint-action conflicts are resolved using an independently resampled priority order $\rho_t$ drawn uniformly over permutations of $\mathcal V$ at every step $t$, as detailed in Section~\ref{subsec:rl_masking}.

Conditioned on the randomized priority order $\rho_t$, the policy and sequential masking mechanism induce a probability distribution over the feasible joint actions,
\begin{equation}\scalebox{0.9}{$
    \pi_t(\cdot\mid\mathcal M_t,\rho_t)
\in
\mathcal P(\mathcal A_{\mathrm{feas}}(\mathbf X_t)),$}
\end{equation}
where $\mathcal P(\cdot)$ denotes the probability simplex over a finite set. Let $\pi:=(\pi_0,\ldots,\pi_{H-1})$ denote a finite-horizon policy, and let $\Pi$ denote the class of admissible policies. After executing $\mathbf a_t$, the robots acquire the measurement batch $\mathcal B_{t+1}$, update the field belief, and receive the reward $R_t$. During training, each episode begins with a new field realization $\tilde f:\mathcal X_{\mathrm g}\rightarrow[f_{\min},f_{\max}]$ generated by the selected stochastic field model. We write $\tilde f\sim\mathbb P_f$, where $\mathbb P_f$ denotes the distribution over field realizations induced by the randomized parameters of that generator defined in Section~\ref{sec:numerical_experiments}. The objective is to learn a policy that maximizes the expected cumulative reward over the finite horizon:
\begin{equation}
\begin{aligned}
\pi^\star
&\in
\arg\max_{\pi\in\Pi} J(\pi),\\
J(\pi)
&:=
\mathbb E^{\pi}_{%
\tilde f\sim\mathbb P_f,\,
\mathbf X_0\sim\mathbb P_0,\,
\{\varepsilon_{i,t}\}_{i\in\mathcal V,\,t=0,\ldots,H},\,
\rho_{0:H-1}}
\left[
\sum_{t=0}^{H-1} R_t
\right],
\end{aligned}
\label{eq:optimal_policy_objective}
\end{equation}
where $\tilde f\sim\mathbb P_f$ denotes a field realization drawn from the training distribution, while $f$ denotes the fixed and unknown field, and $\rho_{0:H-1}:=(\rho_0,\ldots,\rho_{H-1})$. 
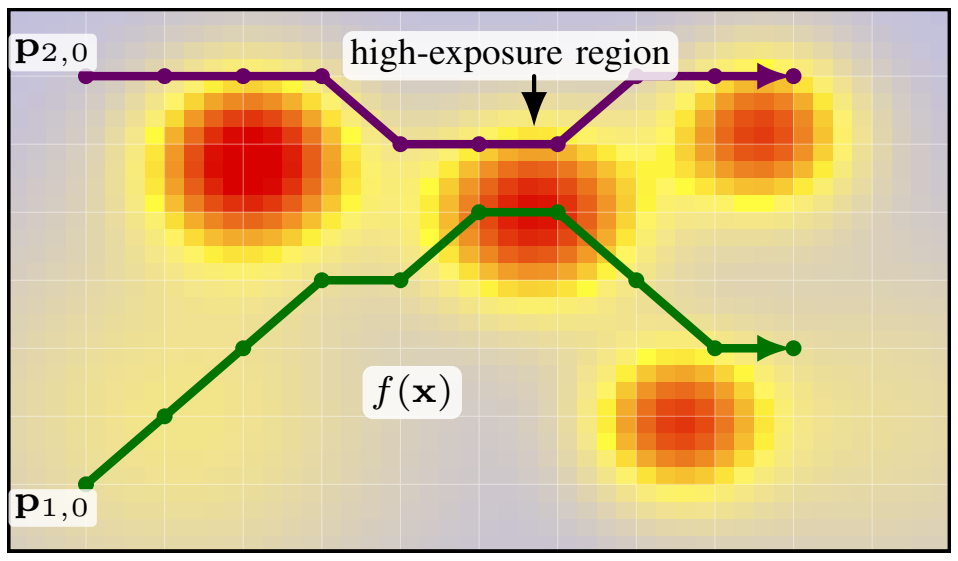
\begin{figure}[t]
\centering
\begin{tikzpicture}[
    x=0.52cm, y=0.45cm, >=Latex,
    line cap=round, line join=round, font=\scriptsize
]
\def\W{12}
\def\H{8}
\colorlet{clow}{blue!22}
\colorlet{cmid}{yellow!75}
\colorlet{chigh}{red!85!black}
\begin{scope}
\clip (0,0) rectangle (\W,\H);
\foreach \i in {0,...,47}{
  \foreach \j in {0,...,31}{
    \pgfmathsetmacro{\xc}{(\i+0.5)*0.25}
    \pgfmathsetmacro{\yc}{(\j+0.5)*0.25}
    \pgfmathsetmacro{\v}{%
        1.00*exp(-(((\xc-3.1)*(\xc-3.1))+((\yc-5.8)*(\yc-5.8)))/(2*1.00*1.00))
      + 0.95*exp(-(((\xc-6.7)*(\xc-6.7))+((\yc-5.0)*(\yc-5.0)))/(2*1.05*1.05))
      + 0.85*exp(-(((\xc-9.6)*(\xc-9.6))+((\yc-6.2)*(\yc-6.2)))/(2*0.90*0.90))
      + 0.75*exp(-(((\xc-8.5)*(\xc-8.5))+((\yc-1.9)*(\yc-1.9)))/(2*0.85*0.85))
      + 0.38*exp(-(((\xc-2.2)*(\xc-2.2))+((\yc-2.0)*(\yc-2.0)))/(2*2.20*2.20))
      + 0.30*exp(-(((\xc-11.0)*(\xc-11.0))+((\yc-2.0)*(\yc-2.0)))/(2*2.00*2.00))}
    \pgfmathsetmacro{\vv}{min(max(\v,0),1)}
    \pgfmathtruncatemacro{\lowmix}{100*min(\vv/0.5,1)}
    \pgfmathtruncatemacro{\highmix}{100*max(0,(\vv-0.5)/0.5)}
    \colorlet{fc}{cmid!\lowmix!clow}
    \colorlet{fc}{chigh!\highmix!fc}
    \fill[fc] (\i*0.25,\j*0.25) rectangle ++(0.26,0.26);
  }
}
\end{scope}
\clip (0,0) rectangle (\W,\H);
\draw[step=1, white, very thin, opacity=0.45] (0,0) grid (\W,\H);
\draw[very thick] (0,0) rectangle (\W,\H);
\draw[green!45!black, line width=1.6pt, -{Latex[length=2.4mm]}]
    (1,1) -- (2,2) -- (3,3) -- (4,4) -- (5,4) -- (6,5) --
    (7,5) -- (8,4) -- (9,3) -- (10,3);
\foreach \p in {(1,1),(2,2),(3,3),(4,4),(5,4),(6,5),(7,5),(8,4),(9,3),(10,3)}
\fill[green!45!black] \p circle (1.5pt);
\draw[violet!80!black, line width=1.6pt, -{Latex[length=2.4mm]}]
    (1,7) -- (2,7) -- (3,7) -- (4,7) -- (5,6) -- (6,6) --
    (7,6) -- (8,7) -- (9,7) -- (10,7);
\foreach \p in {(1,7),(2,7),(3,7),(4,7),(5,6),(6,6),(7,6),(8,7),(9,7),(10,7)}
    \fill[violet!80!black] \p circle (1.5pt);
\node[below left=1pt, fill=white, fill opacity=0.8, text opacity=1,
      inner sep=1pt, rounded corners=1pt] at (1.2,1) {$\mathbf{p}_{1,0}$};
\node[above left=1pt, fill=white, fill opacity=0.8, text opacity=1,
      inner sep=1pt, rounded corners=1pt] at (1.2,7) {$\mathbf{p}_{2,0}$};
\node[align=center, fill=white, fill opacity=0.84, text opacity=1,
      inner sep=1.5pt, rounded corners=2pt] at (5.15,2.35)
      {$f(\mathbf{x})$};
\node[align=center, fill=white, fill opacity=0.84, text opacity=1,
      inner sep=1.5pt, rounded corners=2pt] at (6.4,7.3)
      {high-exposure region};
\draw[black, thick, ->] (6.7,7.0) -- (6.7,6.25);
\begin{scope}[shift={(12.35,1.0)}]
    \shade[bottom color=clow, middle color=cmid, top color=chigh]
        (0,-1) rectangle (0.38,7);
    \draw[thin] (0,-1) rectangle (0.38,7);
    \node[anchor=east] at (1.25,-0.7) {\tiny low};
    \node[anchor=east] at (1.25,6.7) {\tiny high};
\end{scope}
\end{tikzpicture}\vspace{-0.5em}
\caption{Illustration of two sensing agents traversing a scalar field. The background color indicates field intensity, from low (blue) to high (red). Trajectories show the sequential motion of the agents from initial positions $\mathbf{p}_{1,0}$ and $\mathbf{p}_{2,0}$; markers indicate the measurement locations.}
\label{fig:fig1}
\end{figure}

\section{Methodology}
\label{sec:method}
The measurement history $\mathcal D_t$ induces a GP posterior over $\mathcal X_{\mathrm g}$, whose mean and variance summarize the field estimate, while the posterior covariance determines the information objective.

\subsection{Gaussian-Process Model}
\label{subsec:global_gp}
The unknown scalar field is modeled by the GP prior
\begin{equation}
f
\sim
\mathcal{GP}
\left(
\mu_0,\,
k(\mathbf x,\mathbf x')
\right),
\label{eq:global_gp_prior}
\end{equation}
where $\mathbf x,\mathbf x'\in\mathcal{X}_{\mathrm{g}}$ denote spatial locations, with constant prior mean $\mu_0$ and isotropic squared-exponential kernel $k(\mathbf x,\mathbf x') = \sigma_f^2 \exp\!\left( -\frac{ \|\mathbf x-\mathbf x'\|_2^2}{ 2\ell^2 } \right)$, where $\sigma_f^2>0$ is the signal variance and $\ell>0$ is the length scale.

Given $n_t:=|\mathcal D_t|=N(t+1)$, let $Z_t\in\mathbb R^{n_t\times2}$ denote the matrix containing the measurement locations and let $\mathbf y_t\in\mathbb R^{n_t}$ denote the vector of measurements contained in $\mathcal D_t$. For matrix arguments, $K(\cdot,\cdot)$ denotes the kernel matrix whose $(i,j)$ entry is the kernel evaluated between the $i$th row of the first matrix and the $j$th row of the second. Define the regularized Gram matrix $ K_t := K(Z_t,Z_t) + \sigma_n^2 \mathbf I_{n_t}$. Let $\mathbf X_{\mathrm g}\in\mathbb R^{n_c\times2}$ denote the matrix of grid locations in $\mathcal X_{\mathrm g}$, and define $K_{\mathrm{gg}} := K(\mathbf X_{\mathrm g},\mathbf X_{\mathrm g})$, $K_{\mathrm{gt}} := K(\mathbf X_{\mathrm g},Z_t)$.

The posterior mean $\boldsymbol{\mu}_t \in\mathbb R^{n_c}$ over $\mathcal X_{\mathrm g}$ is
\begin{equation}
\boldsymbol{\mu}_t
=
\mu_0\mathbf 1_{n_c}
+
K_{\mathrm{gt}}
K_t^{-1}
\left(
\mathbf y_t-\mu_0\mathbf 1_{n_t}
\right),
\label{eq:global_posterior_mean}
\end{equation}
and the posterior covariance matrix $\Sigma_t \in\mathbb S_{++}^{n_c}$ is
\begin{equation}
\Sigma_t
=
K_{\mathrm{gg}}
-
K_{\mathrm{gt}}
K_t^{-1}
K_{\mathrm{gt}}^{\top}.
\label{eq:global_posterior_covariance_matrix}
\end{equation}
For distinct grid locations, the squared-exponential kernel yields $K_{\mathrm{gg}}\succ0$, which, together with $\sigma_n^2>0$ implies that $\Sigma_t\succ0$ for every finite $t$, so the log-determinant criterion used in Section~\ref{sec:reward} is well defined. The corresponding standard-deviation vector is $\boldsymbol{\sigma}_t := \sqrt{\operatorname{diag}(\Sigma_t)}
\in\mathbb R_{\geq0}^{n_c}$. Let $n_x$ and $n_y$ denote the dimensions of the grid, and let $\mathcal R:\mathbb R^{n_c}\rightarrow\mathbb R^{n_x\times n_y}$ denote the fixed embedding that places values indexed by $\mathcal X_{\mathrm g}$ into their corresponding cells of the rectangular grid and assigns a fixed value $c_{\mathrm{mask}}\in\mathbb R$ to all non-navigable cells. Define the corresponding GP belief maps $M_t:=\mathcal R(\boldsymbol{\mu}_t)$ and $S_t:=\mathcal R(\boldsymbol{\sigma}_t)$.

\subsection{Policy Observation}
\label{subsec:policy_observation}
The posterior mean and standard-deviation maps are supplied to every agent. For agent $i\in\mathcal V$, let $\Phi_{i,t}\in\{0,1\}^{n_x\times n_y}$ denote the binary map marking the grid cell currently occupied by agent $i$, and let $\Phi_{-i,t}\in\{0,1\}^{n_x\times n_y}$ mark the cells currently occupied by all other agents in $\mathcal V\setminus\{i\}$. The navigation map $M^{\mathrm{nav}}\in\{0,1\}^{n_x\times n_y}$ is one if the corresponding cell is navigable and zero otherwise. The remaining horizon is $\tau_t := (H-t)/H\in[0,1]$, $t=0,\ldots,H$, and is replicated over the grid as $T_t := \tau_t\mathbf 1_{n_x\times n_y}\in\mathbb R^{n_x\times n_y}$, thereby providing the fraction of remaining measurement steps at every grid cell. The observation supplied to agent $i$ is obtained by stacking the six $n_x\times n_y$ channels,
\begin{equation}
o_{i,t}
=
\left[
M_t,
S_t,
\Phi_{-i,t},
\Phi_{i,t},
M^{\mathrm{nav}},
T_t
\right].
\label{eq:observation_tensor}
\end{equation}
The channels $(M_t,S_t,M^{\mathrm{nav}},T_t)$ are common to all agents, whereas $\Phi_{i,t}$ and $\Phi_{-i,t}$ are agent-specific. Here $S_t$ identifies regions where sensing can reduce GP uncertainty, whereas $M_t$ provides the estimate of field intensity, and hence indicates regions likely to incur greater exposure. The occupancy channels encode the fleet configuration for spatial allocation, and $T_t$ makes the remaining budget observable to the policy. 

\section{Reward Design}
\label{sec:reward}
The learning objective balances information acquisition against field exposure through the fleet-level scalar reward
\begin{equation}
R_t
=
(1-\lambda)I_t
-
\frac{\lambda}{H}C_t,
\qquad
\lambda\in[0,1],
\label{eq:reward_definition}
\end{equation}
where $I_t$ is the information gained from the fleet measurement batch $\mathcal B_{t+1}$ and $C_t\in[0,1]$ is the fleet-mean exposure. 

\subsection{Fleet Exposure Cost}
\label{subsec:exposure_reward}
During training, the nominal field intensity upper bound $f_{\max}>0$ provides a fixed scale for normalizing the measured field intensity. Exposure is computed from the measurements: positive values contribute to the cost, while negative values arising from measurement noise do not.

For agent $i$, define the normalized field-exposure cost as
\begin{equation}
c_{i,t}
:=
\min\left\{
1,\,
\frac{[y_{i,t+1}]_+}{f_{\max}}
\right\},
\qquad
[z]_+:=\max\{z,0\}.
\label{eq:individual_exposure_cost}
\end{equation}The mean exposure cost for the entire fleet is $ C_t := \frac{1}{N} \sum_{i\in\mathcal V}c_{i,t}
\in[0,1]$. Measurement noise or an unseen field realization may yield $y_{i,t+1}>f_{\max}$; the clipping in \eqref{eq:individual_exposure_cost} assigns such realizations a unit exposure cost to preserve $c_{i,t}\in[0,1]$. 

\subsection{Episode-Level Interpretation}
\label{subsec:episode_reward}
The information gain is denoted by $I_t=I_t^{\mathrm A}$ under A-optimality and $I_t=I_t^{\mathrm D}$ under D-optimality. For A-optimality, define $\mathcal U_{\mathrm A}(\Sigma) :=\operatorname{tr}\Sigma/
\operatorname{tr}\Sigma_{\mathrm{prior}}$ and $I_t^{\mathrm A} :=\mathcal U_{\mathrm A}(\Sigma_t)
-\mathcal U_{\mathrm A}(\Sigma_{t+1})$. The normalized information gains satisfy
\begin{equation}
0
\le
\sum_{t=0}^{H-1}I_t^{\mathrm A}
=
\frac{
\operatorname{tr}\Sigma_0-\operatorname{tr}\Sigma_H
}{
\operatorname{tr}\Sigma_{\mathrm{prior}}
}
\le1.
\label{eq:a_optimal_telescoping}
\end{equation}
For D-optimality, define the dimensionless covariance $\bar{\Sigma}_t:=\Sigma_t/\sigma_f^2$. Under the Gaussian measurement model with $\sigma_n^2>0$, a single measurement yields a log-determinant reduction of $\log(1+\sigma_f^2/\sigma_n^2)$, since the GP posterior marginal variance is bounded by its prior value $\sigma_f^2$~\cite{Qureshi2026HoughTransform}. Sequentially assimilating the $N$ measurements in a batch therefore yields a total log-determinant reduction of at most $N\log(1+\sigma_f^2/\sigma_n^2)$~\cite{rasmussen2006gaussian}. Since the episode contains $H$ such batches, define $ Z_D := NH \log\left( 1+\frac{\sigma_f^2}{\sigma_n^2}
\right)$. The normalized D-optimality reward is then defined as $ I_t^{\mathrm D} := \frac{ \log\det\bar{\Sigma}_t - \log\det\bar{\Sigma}_{t+1} }{ Z_D }$. Accordingly,
\begin{equation}
0
\le
\sum_{t=0}^{H-1}I_t^{\mathrm D}
=
\frac{
\log\det\bar{\Sigma}_0
-
\log\det\bar{\Sigma}_H
}{
Z_D
}
\le 1.
\label{eq:d_optimal_telescoping}
\end{equation}
Summing \eqref{eq:reward_definition} over the episode gives
\begin{equation}
\sum_{t=0}^{H-1}R_t = (1-\lambda)
\sum_{t=0}^{H-1}I_t
-
\frac{\lambda}{H}
\sum_{t=0}^{H-1}C_t,
\label{eq:cumulative_reward}
\end{equation}
\begin{equation}
J(\pi) = \mathbb E^\pi \left[ (1-\lambda) \sum_{t=0}^{H-1}I_t - \frac{\lambda}{H}
\sum_{t=0}^{H-1}C_t \right].
\label{eq:episode_scalarization}
\end{equation} Since $0\le\sum_{t=0}^{H-1}I_t\le1$ and $0\le H^{-1}\sum_{t=0}^{H-1}C_t\le1$, the cumulative reward satisfies $\sum_{t=0}^{H-1}R_t \in [-\lambda,\,1-\lambda]$. Define
\[
\bar I^\pi
:=
\mathbb E^\pi
\left[
\sum_{t=0}^{H-1}I_t
\right],
\qquad
\bar C^\pi
:=
\mathbb E^\pi
\left[
\frac{1}{H}
\sum_{t=0}^{H-1}C_t
\right].
\]
Then $J(\pi) = (1-\lambda)\bar I^\pi - \lambda\bar C^\pi$. Because both $I_t$ and $C_t$ lie in the interval $[0,1]$, $\lambda$ allows the user to hedge between the information-maximization and exposure-minimization objectives. For $0<\lambda<1$, any global maximizer of the scalarized objective lies on the Pareto front corresponding to the objectives of maximizing $\bar I^\pi$ and minimizing $\bar C^\pi$.

\section{Deep Reinforcement-Learning Policy}
\label{sec:rl}
The objective in \eqref{eq:episode_scalarization} is optimized using PPO~\cite{schulman2017ppo}. Homogeneous agents share the actor parameters $\theta_\pi$, while a centralized critic with parameters $\omega$ evaluates the fleet-level state. Each agent uses the observation $o_{i,t}$ in \eqref{eq:observation_tensor} to select its action and receives the common fleet reward $R_t$ in \eqref{eq:reward_definition}.

The shared actor $\pi_{\theta_\pi}(a\mid o_{i,t})$ defines the policy of agent $i$. Define the centralized critic observation as
\begin{equation}
o_t^{\mathrm c} := [ M_t, S_t, \sum_{j\in\mathcal V}\Phi_{j,t}, M^{\mathrm{nav}}, T_t],
\label{eq:critic_observation}
\end{equation}
and let $V_\omega(o_t^{\mathrm c})$ denote the corresponding fleet-value estimate. Thus, execution is decentralized, while learning is centralized. The undiscounted finite-horizon return is optimized. The actor and the centralized critic are parameterized by convolutional neural networks as defined in Section~\ref{sec:experiments}. 

\subsection{Masked Multi-Agent Action Selection}
\label{subsec:rl_masking}
The per-agent network policies described above are converted into a feasible joint action using the randomized sequential masking mechanism introduced in Section~\ref{sec:problem}. At each action step $t$, a priority permutation $\rho_t$ is sampled uniformly over all permutations of $\mathcal V$. Let $\rho_t(k)\in\mathcal V$ denote the agent at priority position $k\in\{1,\ldots,N\}$. Actions are then selected in the order $\rho_t(1),\ldots,\rho_t(N)$.

For agent $i=\rho_t(k)$, let $\mathcal{A}_{i,t}^{\mathrm{mask}}\subseteq\mathcal {A}_{\mathrm{nav},i,t}$ denote the actions remaining after the higher-priority agents $\rho_t(1),\ldots,\rho_t(k-1)$ have sequentially sampled their actions from their masked policy distributions. Agent $i$ samples $a_{i,t}\sim
\pi_{\theta_\pi}^{\mathrm{mask}}\!\left(
\cdot\mid o_{i,t},\mathcal A_{i,t}^{\mathrm{mask}}
\right)$. The navigation-feasible set $\mathcal A_{\mathrm{nav},i,t}$ defined in \eqref{eq:navigation_action_set} enforces boundary, obstacle, and diagonal corner-cutting constraints. An action $a\in\mathcal A_{\mathrm{nav},i,t}$ is removed from $\mathcal A_{i,t}^{\mathrm{mask}}$ if, together with a previously selected action, it would cause two agents to reach the same destination, exchange positions, produce intersecting motion segments, or move into the current position of another agent. Because the wait action remains available and motion into another agent's current position is excluded, $\mathcal A_{i,t}^{\mathrm{mask}}\neq\varnothing$ for all $i\in\mathcal V$ and $t\in\{0,\ldots, H-1\}$. The selected actions then form a joint action $\mathbf a_t\in\mathcal A_{\mathrm{feas}}(\mathbf X_t)$.

For PPO, let $z_{\theta_\pi}(o_{i,t},a)$ denote the score assigned to action $a$ given observation $o_{i,t}$. The masked per-agent policy is
\begin{equation*}\scalebox{0.85}{$
\pi_{\theta_\pi}^{\mathrm{mask}}
\left(
a\mid
o_{i,t},
\mathcal A_{i,t}^{\mathrm{mask}}
\right)
=
\begin{cases}
\displaystyle
\frac{
\exp z_{\theta_\pi}(o_{i,t},a)
}{
\sum\limits_{a'\in\mathcal A_{i,t}^{\mathrm{mask}}}
\exp z_{\theta_\pi}(o_{i,t},a')
},
&
a\in\mathcal A_{i,t}^{\mathrm{mask}},
\\[3mm]
0,
&
a\notin\mathcal A_{i,t}^{\mathrm{mask}}.
\end{cases}$}
\label{eq:masked_policy}
\end{equation*}
Thus, masked actions receive zero probability, while the probabilities of the remaining feasible actions are renormalized over $\mathcal A_{i,t}^{\mathrm{mask}}$.  The sequential composition of these masked per-agent decisions, conditioned on $\rho_t$, induces the admissible joint policy $\pi_t(\cdot\mid\mathcal M_t,\rho_t)$ introduced in Section~\ref{sec:problem}. For each rollout sample, the realized mask $\mathcal A_{i,t}^{\mathrm{mask}}$, generated from the sampled permutation $\rho_t$ and the actions sampled by the higher-priority agents, is stored and reused when evaluating $a_{i,t}$ during the PPO update.

\subsection{Proximal Policy Optimization}
\label{subsec:ppo} PPO trains the shared masked actor and centralized fleet critic using generalized advantage estimation (GAE) and the clipped surrogate objective~\cite{schulman2016gae,schulman2017ppo}. The temporal-difference error is
\begin{equation}
\delta_t
=
R_t
+
\gamma(1-d_t)
V_{\omega_{\mathrm{old}}}(o_{t+1}^{\mathrm c})
-
V_{\omega_{\mathrm{old}}}(o_t^{\mathrm c}),
\label{eq:fleet_td_error}
\end{equation}
where $d_t:=\mathbf 1\{t=H-1\}$ and $\gamma=1$. GAE gives $ \widehat A_t^{\mathrm F} = \delta_t + \lambda_{\mathrm{GAE}} (1-d_t) \widehat A_{t+1}^{\mathrm F}$, with $\widehat A_H^{\mathrm F}:=0$. The advantage $\widehat A_t^{\mathrm F}$ is assigned to all agent actions selected at step $t$, and the corresponding critic target is $\widehat R_t^{\mathrm F} := \widehat A_t^{\mathrm F} + V_{\omega_{\mathrm{old}}}(o_t^{\mathrm c})$, where $V_{\omega_{\mathrm{old}}}(o_t^{\mathrm c})$ is the value estimate stored during rollout collection. Let $\mathcal H(\cdot)$ denote categorical policy entropy, $\operatorname{Huber}(\cdot)$ the Huber loss, $\beta_{\mathrm{ent}}\ge0$ the entropy-regularization coefficient, $\mathbb E_{t,i}$ empirical averaging over rollout steps and agents, and $\mathbb E_t$ empirical averaging over rollout steps.

Let $\theta_{\pi,\mathrm{old}}$ denote the actor parameters used to generate the current rollout. For each agent-step pair, the likelihood ratio is
$\varrho_{i,t} = \frac{
\pi_{\theta_\pi}^{\mathrm{mask}}
\left(
a_{i,t}
\mid
o_{i,t},
\mathcal A_{i,t}^{\mathrm{mask}}
\right)
}{
\pi_{\theta_{\pi,\mathrm{old}}}^{\mathrm{mask}}
\left(
a_{i,t}
\mid
o_{i,t},
\mathcal A_{i,t}^{\mathrm{mask}}
\right)
}$. Let $\epsilon_{\mathrm c}>0$ denote the PPO clipping parameter; the actor loss is defined as
\begin{equation*}
\begin{split}
\mathcal L_{\mathrm{PPO}}^{\mathrm{actor}}(\theta_\pi)
=
&-
\mathbb E_{t,i}
\Bigl[
\min
\Bigl(
\varrho_{i,t}\widehat A_t^{\mathrm F},
\\
&\qquad\qquad
\operatorname{clip}
\left(
\varrho_{i,t},
1-\epsilon_{\mathrm c},
1+\epsilon_{\mathrm c}
\right)
\widehat A_t^{\mathrm F}
\Bigr)
\Bigr]
\\
&-
\beta_{\mathrm{ent}}
\mathbb E_{t,i}
\left[
\mathcal H
\left(
\pi_{\theta_\pi}^{\mathrm{mask}}
\left(
\cdot\mid
o_{i,t},
\mathcal A_{i,t}^{\mathrm{mask}}
\right)
\right)
\right].
\end{split}
\label{eq:ppo_actor_loss}
\end{equation*}
The centralized critic is trained against $\widehat R_t^{\mathrm F}$ using $ \mathcal L_{\mathrm{PPO}}^{\mathrm{critic}}(\omega) = \mathbb E_t \left[
\operatorname{Huber}
\left(
V_\omega(o_t^{\mathrm c})
-
\widehat R_t^{\mathrm F}
\right)
\right]$.

\subsection{Training and Deployment}
\label{subsec:training_deployment}
Each training episode begins with an independently generated field realization $\tilde f\sim\mathbb P_f$ and an initial fleet configuration $\mathbf X_0\sim\mathbb P_0$, consistent with \eqref{eq:optimal_policy_objective}. During training, a field range $[f_{\min},f_{\max}]$ is assumed for field calibration and normalization. At evaluation, the deployed policy retains this training calibration, but the maximum of unseen test fields is assumed to be unknown. The GP is initialized from the fixed prior of Section~\ref{subsec:global_gp} and conditioned on the initial measurement batch $\mathcal B_0$ to obtain the posterior belief at $t=0$.

For a given training configuration, the GP prior mean $\mu_0$, kernel parameters $(\sigma_f^2,\ell)$, measurement-noise variance $\sigma_n^2$, reward trade-off coefficient $\lambda$, field bounds $[f_{\min},f_{\max}]$, and selected A- or D-optimal information criterion remain fixed. The PPO actor and critic parameters are updated across episodes, whereas the GP belief is reinitialized for each sampled field realization.

Deployment only requires the actor and the GP hyperparameters; the critic parameters are needed only for training. During evaluation, the GP posterior is updated online and each agent selects the action with the largest probability under $\pi_{\theta_\pi}^{\mathrm{mask}}$. Actions are selected according to $\rho_t$, yielding $\mathbf a_t\in\mathcal A_{\mathrm{feas}}(\mathbf X_t)$.

\section{Numerical Experiments}
\label{sec:numerical_experiments}
\begin{figure}[t]
\centering
\includegraphics[width=\columnwidth]{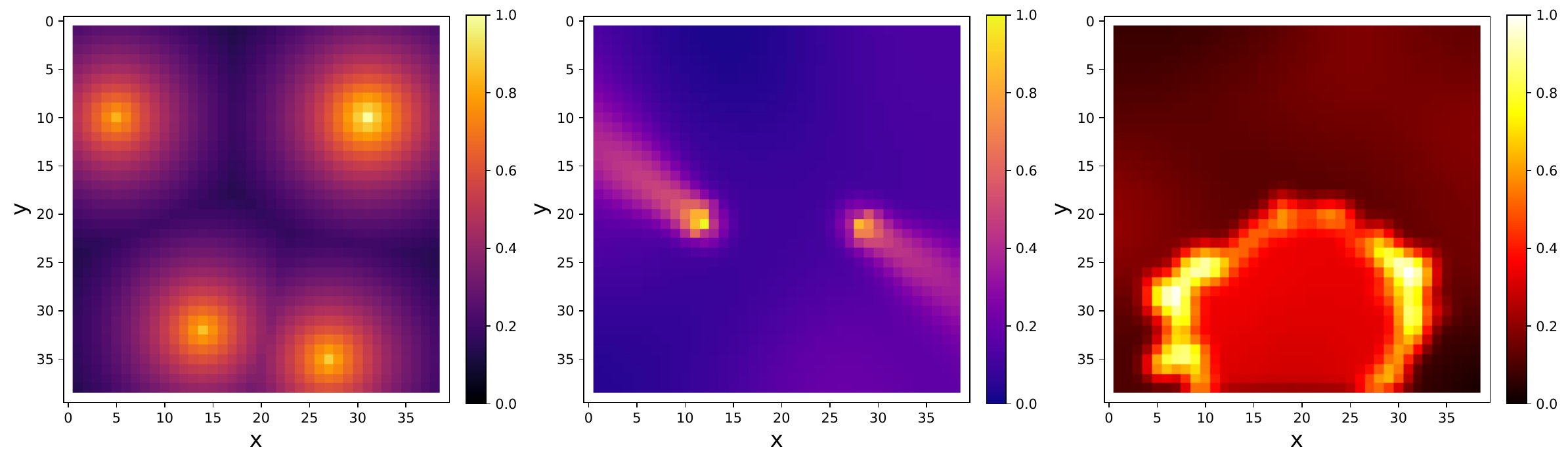}\vspace{-0.5em}
\caption{From left to right: radar, plume, and wildfire-field realizations.}
\label{fig:fig2}
\end{figure}

At the start of each training episode, a generator produces a randomized scalar-valued function $g:\mathcal X_{\mathrm g}\rightarrow[0,1]$, which is clipped between upper and lower bounds to yield the scalar field $f(\mathbf x)=f_{\min}+(f_{\max}-f_{\min})g(\mathbf x)$. To evaluate the developed EA-IPP technique, three field families are considered: radar fields, plumes, and wildfire fronts, as illustrated in Figure~\ref{fig:fig2}.

\subsection{Radar Field}
\label{subsec:radar_family}
Each realization of the radar scalar field contains four stationary emitters, with one location $\mathbf z_q\in\mathcal X_{\mathrm g}$ sampled from each workspace quadrant. Candidate cells exclude a four-cell boundary margin and a three-cell band around the quadrant-dividing lines. Sample $\widetilde A_q\sim\mathcal U(0.8,1)$ for $q\in\{1,\ldots,4\}$, set the largest sampled amplitude to $1$, and let the remaining amplitudes satisfy $A_q=\widetilde A_q$. The field is then given by
\begin{equation}
g_{\mathrm{radar}}(\mathbf x) = 
\max_{q\in\{1,\ldots,4\}}
\left[
A_q
\exp\!\left(
-\frac{\|\mathbf x-\mathbf z_q\|_2}{8\Delta}
\right)
\right],
\label{eq:radar_field}
\end{equation}
Emitter locations and amplitudes are resampled between episodes.

\subsection{Plume Field}
\label{subsec:plume_family}
Each realization in the plume field contains $n_{\mathrm s}=2$ sources. Let $(r(x),c(x))$ denote the grid indices of $\mathbf x\in\mathcal X_{\mathrm g}$. Source locations lie at least five cells from the boundary, with one satisfying $c-r\ge 3$ and the other $r-c\ge 3$. Let $(r_q,c_q)$ denote the grid indices of source $q$. For $q\in\{1,2\}$, the source strengths $A_q$ are sampled independently from $\mathcal U(0.75,1)$, after which the larger of $A_1$ and $A_2$ is set to $1$. Each source independently receives $\theta_q\sim\mathcal U(0,2\pi)$. With $\delta r_q (\mathbf{x})=r (\mathbf{x}) -r_q$ and $\delta c_q (\mathbf{x}) = c(\mathbf{x}) - c_q$, define $u_q (\mathbf{x}) =\delta c_q(\mathbf{x})\cos\theta_q+\delta r_q(\mathbf{x})\sin\theta_q$ and $v_q(\mathbf{x})=-\delta c_q(\mathbf{x})\sin\theta_q+\delta r_q(\mathbf{x})\cos\theta_q$. The plume width is $\omega_q(\mathbf{x}):=1.2+0.16[u_q(\mathbf{x})]_+$. The field is
$\widetilde g_{\mathrm{plume}}(\mathbf x)
=
\sum_{q=1}^{n_{\mathrm s}}
A_q
\Big[
\exp\left(
-\frac{[u_q(\mathbf{x})]_+}{24}
\right)
\exp\left(
-\frac{v_q(\mathbf{x})^2}{2\omega_q^2}
\right)
\mathbf 1\{u_q(\mathbf{x})\ge0\} +
0.85
\exp\left(
-\frac{
\delta r_q^2(\mathbf{x})+\delta c_q^2(\mathbf{x})
}{
21.5^2
}
\right)
\Big].
$ The normalized field is $g_{\mathrm{base}}(\mathbf x)
=
\frac{\widetilde g_{\mathrm{plume}}(\mathbf x)}
{\max_{\mathbf x'\in\mathcal X_{\mathrm g}}
\widetilde g_{\mathrm{plume}}(\mathbf x')}$. An independent i.i.d. standard Gaussian field is smoothed with standard deviation of $7$. Let $z(\mathbf x)$ denote the resulting field and let $b(\mathbf x)=0.10+0.9\frac{z(\mathbf x)-z_{\min}}{z_{\max}-z_{\min}}$, where $z_{\min}$ and $z_{\max}$ are the minimum and maximum of $z$ over $\mathcal X_{\mathrm g}$. The plume field is generated as
\[
g_{\mathrm{plume}}(\mathbf x)
=
\operatorname{clip}_{[0,1]}
\left(
g_{\mathrm{base}}(\mathbf x)
+
0.20
\bigl(1-g_{\mathrm{base}}(\mathbf x)\bigr)b(\mathbf x)
\right).
\]

\subsection{Wildfire}
\label{subsec:fire_family}
The wildfire family is generated by a stochastic cellular automaton and frozen before sensing begins. Let $\mathcal C_\nu$ denote the burning cells, $F_\nu(\mathbf x)$ the remaining fuel, and $\mathcal B_\nu^{\mathrm h}$ the set of previously burned cells at automaton step $\nu$. Initially, $F_0(\mathbf x)=10$ for $\mathbf x\in\mathcal X_{\mathrm g}$; one or two ignition locations are drawn independently and uniformly from $\mathcal X_{\mathrm g}$ with replacement.

A wind direction $\phi\sim\mathcal U(0,2\pi)$ and magnitude $\varsigma_{\mathrm w}\sim\mathcal U(0,0.1)$ define the wind speed $\mathbf w=\varsigma_{\mathrm w}[\cos\phi,\sin\phi]^\top$ in row-column coordinates. For a neighboring grid displacement $\boldsymbol\delta$ satisfying $0<\|\boldsymbol\delta\|_2\le2$, the ignition probability is
$p(\boldsymbol\delta) = \min\!\left\{
1,\, \alpha_{\mathrm f}
\frac{ 1+\boldsymbol\delta^\top\mathbf w
}{ \|\boldsymbol\delta\|_2^2 }
\right\},$ with $\alpha_{\mathrm f}=10^{-2}$. The implementation combines multiple ignition attempts multiplicatively through their non-ignition probabilities and decreases the fuel of burning cells at each update. The $\nu$-step wildfire is then given by
\begin{equation}
\widetilde g_{\mathrm{fire},\nu}(\mathbf x)
=
\min\!\left\{
1,\,
\left[
G\ast
\left(
\mathbf 1_{\mathcal C_\nu}
+
\beta_{\mathrm h}\mathbf 1_{\mathcal B_\nu^{\mathrm h}}
\right)
\right](\mathbf x)
\right\},
\label{eq:fire_field_dynamic}
\end{equation}
where $\beta_{\mathrm h}=0.25$, $G$ is the normalized $5\times5$ Gaussian smoothing kernel, and $\ast$ denotes discrete two-dimensional convolution. Letting $\epsilon_f(\mathbf x)\overset{\mathrm{i.i.d.}}{\sim}\mathcal N(0,1)$ and $z(\mathbf x)=(G_{\sigma_{\mathrm b}}\ast\epsilon_f)(\mathbf x)$, where $G_{\sigma_{\mathrm b}}$ is a Gaussian smoothing kernel with $\sigma_{\mathrm b}=6$. Its rescaled form is $b(\mathbf x)=0.15+0.85\frac{z(\mathbf x)-\min_{\mathbf x'\in\mathcal X_{\mathrm g}}z(\mathbf x')}{\max_{\mathbf x'\in\mathcal X_{\mathrm g}}z(\mathbf x')-\min_{\mathbf x'\in\mathcal X_{\mathrm g}}z(\mathbf x')}$. The resulting field is
\begin{equation}\scalebox{0.9}{$
g_{\mathrm{fire},\nu}(\mathbf x)
=
\operatorname{clip}_{[0,1]}
\left(
\widetilde g_{\mathrm{fire},\nu}(\mathbf x)
+
0.20\,m_\nu
\left(
1-\frac{\widetilde g_{\mathrm{fire},\nu}(\mathbf x)}{m_\nu}
\right)
b(\mathbf x)
\right)$},
\end{equation} where $m_\nu := \max_{\mathbf x\in\mathcal X_{\mathrm g}} \widetilde g_{\mathrm{fire},\nu}(\mathbf x)$. The experiments use the field after 80 steps, i.e., $g_{\mathrm{fire}}(\mathbf x):=g_{\mathrm{fire},80}(\mathbf x)$.

\section{Experimental Setup}
\label{sec:experiments}

\def\PPOChemicalPlumeDLZeroSeed0{data11/learned/ppo_chemical_plume_d_lambda0_seed0}
\def\PPOChemicalPlumeDLZero{data11/learned/ppo_chemical_plume_d_lambda0_seed0}
\def\PPOChemicalPlumeDLQuarterSeed0{data11/learned/ppo_chemical_plume_d_lambda0.25_seed0}
\def\PPOChemicalPlumeDLQuarter{data11/learned/ppo_chemical_plume_d_lambda0.25_seed0}
\def\PPOChemicalPlumeDLHalfSeed0{data11/learned/ppo_chemical_plume_d_lambda0.5_seed0}
\def\PPOChemicalPlumeDLHalf{data11/learned/ppo_chemical_plume_d_lambda0.5_seed0}
\def\PPOChemicalPlumeDLThreeQuarterSeed0{data11/learned/ppo_chemical_plume_d_lambda0.75_seed0}
\def\PPOChemicalPlumeDLThreeQuarter{data11/learned/ppo_chemical_plume_d_lambda0.75_seed0}
\def\PPOChemicalPlumeDLOneSeed0{data11/learned/ppo_chemical_plume_d_lambda1_seed0}
\def\PPOChemicalPlumeDLOne{data11/learned/ppo_chemical_plume_d_lambda1_seed0}
\def\PPORadarALZeroSeed0{data11/learned/ppo_radar_a_lambda0_seed0}
\def\PPORadarALZero{data11/learned/ppo_radar_a_lambda0_seed0}
\def\PPORadarALQuarterSeed0{data11/learned/ppo_radar_a_lambda0.25_seed0}
\def\PPORadarALQuarter{data11/learned/ppo_radar_a_lambda0.25_seed0}
\def\PPORadarALHalfSeed0{data11/learned/ppo_radar_a_lambda0.5_seed0}
\def\PPORadarALHalf{data11/learned/ppo_radar_a_lambda0.5_seed0}
\def\PPORadarALThreeQuarterSeed0{data11/learned/ppo_radar_a_lambda0.75_seed0}
\def\PPORadarALThreeQuarter{data11/learned/ppo_radar_a_lambda0.75_seed0}
\def\PPORadarALOneSeed0{data11/learned/ppo_radar_a_lambda1_seed0}
\def\PPORadarALOne{data11/learned/ppo_radar_a_lambda1_seed0}
\def\PPORadarDLZeroSeed0{data11/learned/ppo_radar_d_lambda0_seed0}
\def\PPORadarDLZero{data11/learned/ppo_radar_d_lambda0_seed0}
\def\PPORadarDLQuarterSeed0{data11/learned/ppo_radar_d_lambda0.25_seed0}
\def\PPORadarDLQuarter{data11/learned/ppo_radar_d_lambda0.25_seed0}
\def\PPORadarDLHalfSeed0{data11/learned/ppo_radar_d_lambda0.5_seed0}
\def\PPORadarDLHalf{data11/learned/ppo_radar_d_lambda0.5_seed0}
\def\PPORadarDLThreeQuarterSeed0{data11/learned/ppo_radar_d_lambda0.75_seed0}
\def\PPORadarDLThreeQuarter{data11/learned/ppo_radar_d_lambda0.75_seed0}
\def\PPORadarDLOneSeed0{data11/learned/ppo_radar_d_lambda1_seed0}
\def\PPORadarDLOne{data11/learned/ppo_radar_d_lambda1_seed0}
\def\PPOWildfireDLZeroSeed0{data11/learned/ppo_wildfires_d_lambda0_seed0}
\def\PPOWildfireDLZero{data11/learned/ppo_wildfires_d_lambda0_seed0}
\def\PPOWildfireDLQuarterSeed0{data11/learned/ppo_wildfires_d_lambda0.25_seed0}
\def\PPOWildfireDLQuarter{data11/learned/ppo_wildfires_d_lambda0.25_seed0}
\def\PPOWildfireDLHalfSeed0{data11/learned/ppo_wildfires_d_lambda0.5_seed0}
\def\PPOWildfireDLHalf{data11/learned/ppo_wildfires_d_lambda0.5_seed0}
\def\PPOWildfireDLThreeQuarterSeed0{data11/learned/ppo_wildfires_d_lambda0.75_seed0}
\def\PPOWildfireDLThreeQuarter{data11/learned/ppo_wildfires_d_lambda0.75_seed0}
\def\PPOWildfireDLOneSeed0{data11/learned/ppo_wildfires_d_lambda1_seed0}
\def\PPOWildfireDLOne{data11/learned/ppo_wildfires_d_lambda1_seed0}
\def\GreedyDChemicalPlume{data11/baselines/greedy_d_chemical_plume}
\def\LawnMowerChemicalPlume{data11/baselines/lawn_mower_chemical_plume}
\def\MaxVarChemicalPlume{data11/baselines/max_variance_chemical_plume}
\def\MinExposureChemicalPlume{data11/baselines/min_exposure_chemical_plume}
\def\RandomChemicalPlume{data11/baselines/random_chemical_plume}
\def\GreedyDRadar{data11/baselines/greedy_d_radar}
\def\LawnMowerRadar{data11/baselines/lawn_mower_radar}
\def\MaxVarRadar{data11/baselines/max_variance_radar}
\def\MinExposureRadar{data11/baselines/min_exposure_radar}
\def\RandomRadar{data11/baselines/random_radar}
\def\GreedyDWildfire{data11/baselines/greedy_d_wildfires}
\def\LawnMowerWildfire{data11/baselines/lawn_mower_wildfires}
\def\MaxVarWildfire{data11/baselines/max_variance_wildfires}
\def\MinExposureWildfire{data11/baselines/min_exposure_wildfires}
\def\RandomWildfire{data11/baselines/random_wildfires}
\def\PaperDataRoot{data11}
\def\PaperTables{data11/tables}
\def\TrainingTables{data11/training}

Motivated by CNN-based architectures previously used for spatial information gathering~\cite{SCC.YanesLuis.Shutin.ea2024}, the shared actor and centralized critic each use three $3\times3$ convolutional layers with $64$, $32$, and $16$ output channels, followed by ReLU activations and a $512$-dimensional feature projection. Separate fully connected trunks with dimensions $512\to256\to128$ are used for the actor and critic.

\textbf{Evaluation matrix.}
PPO is evaluated over $\lambda=\{0,0.25,0.5,0.75,1\}$. The radar field uses both A- and D-optimal information rewards, whereas the plume and wildfire fields use D-optimality, yielding 20 trained policies.

\textbf{Configuration.}
Experiments use a $40\times40$ grid with $n_c=1444$ navigable cells, $N=3$ agents, horizon $H=50$, training-field range $[f_{\min},f_{\max}]=[0,100]$, and measurement-noise standard deviation $\sigma_n=1$. The upper bound $f_{\max}$ is used for GP calibration and exposure normalization. The fields generated during evaluation attain maximum values ranging between \(90\%\) and \(100\%\) of \(f_{\max}\). The true maximum of test fields is not exposed to the policy, and the reward definition and its normalization remain unchanged from training. Since $f_{\max}^{test}\le f_{\max}$, the training bound remains conservative, and the reward function remains well-defined. Table~\ref{tab:table1} lists the GP and PPO settings.

\begin{table}[t]
\centering
\caption{Experimental parameters.}
\label{tab:table1}
\vspace{-1em}
\scriptsize
\setlength{\tabcolsep}{4.5pt}
\renewcommand{\arraystretch}{0.5}
\begin{tabular}{lll}
\toprule
Category & Parameter & Value \\
\midrule
Environment
& Grid / navigable cells & $40\times40$ / $1444$ \\
& $N$ / $H$ & $3$ / $50$ \\
\\
& $\sigma_n$ / $|\mathcal A|$ & $1$ / $9$ \\
\midrule
Global GP
& $\mu_0$ (radar/fire/plume) & $36.31/18.86/6.34$ \\
& $\ell$ (radar/fire/plume) & $4.1891/2.0738/2.1928$ \\
& $\sigma_f$ (radar/fire/plume) & $13.335/19.806/9.6393$ \\
\midrule
PPO
& Discount factor $\gamma$ & $1$ \\
& Feature dimension & $512$ \\
& Learning rate & $5\times10^{-5}$ \\
& $\lambda_{\mathrm{GAE}}$ / clip ratio $\epsilon_{\mathrm c}$ & $0.95$ / $0.2$ \\
& Value / entropy coefficient & $0.35$ / $0.01$ \\
& Update epochs / minibatch size & $4$ / $128$ \\
& Episodes per update / gradient clip & $8$ / $0.5$ \\
\bottomrule
\end{tabular}\vspace{-2em}
\end{table}

\textbf{Comparison baselines.}
All baselines use the same motion constraints, feasible-action masking, GP belief, and randomized sequential agent-priority rule as PPO; only the action-selection rule changes. \emph{Greedy-D} selects
\begin{equation}\scalebox{0.9}{$
a_{i,t}^{\mathrm{G}\text{-}\mathrm D}
=
\arg\max_{a\in\mathcal A_{i,t}^{\mathrm{feas}}}
I_t^{\mathrm D}(a),$}
\end{equation}
where $I_t^{\mathrm D}(a)$ is the previously defined one-step D-optimal information gain evaluated for candidate action $a$~\cite{krause2008near}. \emph{MaxVar} selects
\begin{equation}\scalebox{0.9}{$
a_{i,t}^{\mathrm{MV}}
=
\arg\max_{a\in\mathcal A_{i,t}^{\mathrm{feas}}}
\sigma_t^2(\mathbf x_{i,t}^{a}),$}
\end{equation}
corresponding to maximum variance-driven GP sampling~\cite{krause2008near}. \emph{MinExposure} selects
\begin{equation}\scalebox{0.9}{$
a_{i,t}^{\mathrm{ME}}
=
\arg\min_{a\in\mathcal A_{i,t}^{\mathrm{feas}}}
\widehat c_{i,t}(a),$}
\end{equation}
where $\widehat c_{i,t}(a)
=
\min\!\left\{
1,\,
\frac{[\mu_t(\mathbf x_{i,t}^{a})]_+}{f_{\max}}
\right\}$ is the GP-mean prediction of the normalized exposure in \eqref{eq:individual_exposure_cost}. \emph{Lawn-mower} follows a fixed coverage order, while \emph{Random} samples uniformly from $\mathcal A_{i,t}^{\mathrm{feas}}$~\cite{popovic2024survey}. Together, the baselines span information-driven, exposure-driven, coverage-driven, and uninformed sensing strategies, but none optimize the information--exposure trade-off offered by the EA-IPP policy. SafeOpt-style methods impose safety constraints through a safety threshold and a certified safe set and are therefore not directly comparable to the penalized exposure objective considered here~\cite{sui2015safe,turchetta2016safe}.

\textbf{Evaluation metrics.}
For $q\in\{A,D\}$, the episode information return is $ I_{\mathrm{ep}}^{q} := \sum_{t=0}^{H-1} I_t^{q}$. The corresponding episode-level exposure is $C_{\mathrm{ep}} := \frac{1}{H} \sum_{t=0}^{H-1}C_t$. Root mean square error (RMSE) at step $t$ is computed as
$\operatorname{RMSE}_t :=
\sqrt{
\frac{1}{n_c}
\sum_{j=1}^{n_c}
\left[
\mu_t(\mathbf x^{(j)})
-
f(\mathbf x^{(j)})
\right]^2
}$. The negative log predictive density (NLPD) is
$
\operatorname{NLPD}_t
:=
\frac{1}{n_c}
\sum_{j=1}^{n_c}
\left[
\frac{1}{2}
\log\!\left(
2\pi\sigma_t^2(\mathbf x^{(j)})
\right)
+
\frac{
\left[
f(\mathbf x^{(j)})-\mu_t(\mathbf x^{(j)})
\right]^2
}{
2\sigma_t^2(\mathbf x^{(j)})
}
\right]$. Together, these metrics quantify information acquisition, cumulative exposure, reconstruction accuracy, and predictive calibration, providing a complementary evaluation of the learned policies.

\textbf{Evaluation protocol.} Each policy is evaluated using 1000 held-out episodes, distinct from the training data. Baselines are evaluated on the same randomly generated evaluation fields, initial fleet configurations, measurement-noise realizations, and randomized agent-priority sequences. 

\section{Results}
\label{sec:results}
\subsection{Effect of $\lambda$ on Information, Exposure, and GP error}
\label{subsec:results_lambda}
Figure~\ref{fig:fig3} shows results of $\lambda$-ablation. Across all three field families, larger $\lambda$ values generally correspond to lower information gain and exposure and higher RMSE and NLPD, demonstrating a trade-off between exposure reduction and GP reconstruction performance. For example, in the plume field, $\lambda=0.25$ reduces $C_{\mathrm{ep}}$ from $0.165$ to $0.152$ while retaining $88.3\%$ of the D-optimal information ($0.557$ to $0.492$).
\begin{figure*}
\centering

\def\RadarLambdaCSV{data11/tables/lambda_ablation/radar_d_lambda_ablation.csv}
\def\ChemicalLambdaCSV{data11/tables/lambda_ablation/chemical_plume_d_lambda_ablation.csv}
\def\WildfireLambdaCSV{data11/tables/lambda_ablation/wildfires_d_lambda_ablation.csv}

\begin{tikzpicture}
\begin{groupplot}[
    group style={
        group size=3 by 2,
        horizontal sep=1.1cm,
        vertical sep=1.1cm,
    },
    width=0.36\textwidth,
    height=0.18\textwidth,
    xmin=-0.05,
    xmax=1.05,
    ylabel style={xshift=3pt,yshift=-4pt},
    xlabel style={yshift=6pt},
    xtick={0,0.25,0.5,0.75,1},
    xlabel={$\lambda$},
    grid=both,
    major grid style={line width=.2pt},
    minor grid style={line width=.1pt},
    tick label style={font=\scriptsize},
    label style={font=\scriptsize},
    title style={font=\scriptsize,yshift=-9pt},
    legend style={
        font=\scriptsize,
        draw=none,
        legend columns=3,
        cells={anchor=west},
        /tikz/every even column/.append style={column sep=7pt},
    },
]

\nextgroupplot[
    title={(a) A-optimal information},
    ylabel={$I_{\mathrm{ep}}^{\mathrm A}$},
    legend to name=lambdaFieldLegend,
]

\addplot+[mark=*, thick]
table[x=lambda,y=infoAMean,col sep=comma]{\RadarLambdaCSV};
\addlegendentry{Radar}

\addplot+[mark=square*, thick]
table[x=lambda,y=infoAMean,col sep=comma]{\ChemicalLambdaCSV};
\addlegendentry{Plume}

\addplot+[mark=triangle*, thick]
table[x=lambda,y=infoAMean,col sep=comma]{\WildfireLambdaCSV};
\addlegendentry{Wildfire}

\nextgroupplot[
    title={(b) D-optimal information},
    ylabel={$I_{\mathrm{ep}}^{\mathrm D}$},
]

\addplot+[mark=*, thick]
table[x=lambda,y=infoDMean,col sep=comma]{\RadarLambdaCSV};

\addplot+[mark=square*, thick]
table[x=lambda,y=infoDMean,col sep=comma]{\ChemicalLambdaCSV};

\addplot+[mark=triangle*, thick]
table[x=lambda,y=infoDMean,col sep=comma]{\WildfireLambdaCSV};

\nextgroupplot[
    title={(c) Episode exposure},
    ylabel={$C_{\mathrm{ep}}$},
]

\addplot+[mark=*, thick]
table[x=lambda,y=exposureMean,col sep=comma]{\RadarLambdaCSV};

\addplot+[mark=square*, thick]
table[x=lambda,y=exposureMean,col sep=comma]{\ChemicalLambdaCSV};

\addplot+[mark=triangle*, thick]
table[x=lambda,y=exposureMean,col sep=comma]{\WildfireLambdaCSV};

\nextgroupplot[
    title={(d) Reconstruction error},
    ylabel={RMSE},
]

\addplot+[mark=*, thick]
table[x=lambda,y=rmseMean,col sep=comma]{\RadarLambdaCSV};

\addplot+[mark=square*, thick]
table[x=lambda,y=rmseMean,col sep=comma]{\ChemicalLambdaCSV};

\addplot+[mark=triangle*, thick]
table[x=lambda,y=rmseMean,col sep=comma]{\WildfireLambdaCSV};

\nextgroupplot[
    title={(e) Negative log predictive density},
    ylabel={NLPD},
]

\addplot+[mark=*, thick]
table[x=lambda,y=nlpdMean,col sep=comma]{\RadarLambdaCSV};

\addplot+[mark=square*, thick]
table[x=lambda,y=nlpdMean,col sep=comma]{\ChemicalLambdaCSV};

\addplot+[mark=triangle*, thick]
table[x=lambda,y=nlpdMean,col sep=comma]{\WildfireLambdaCSV};

\nextgroupplot[
    title={(f) Predictive coverage},
    ylabel={$\mathrm{Cov}_{95}$},
]

\addplot+[mark=*, thick]
table[x=lambda,y=coverage95Mean,col sep=comma]{\RadarLambdaCSV};

\addplot+[mark=square*, thick]
table[x=lambda,y=coverage95Mean,col sep=comma]{\ChemicalLambdaCSV};

\addplot+[mark=triangle*, thick]
table[x=lambda,y=coverage95Mean,col sep=comma]{\WildfireLambdaCSV};

\end{groupplot}

\node at ($(group c2r2.south)+(0,-0.73cm)$)
{\ref{lambdaFieldLegend}};

\end{tikzpicture}

\caption{Effect of $\lambda$ on D-optimal PPO across the three field families. Top: $I_{\mathrm{ep}}^{\mathrm A}$, $I_{\mathrm{ep}}^{\mathrm D}$, and $C_{\mathrm{ep}}$. Bottom: terminal RMSE, NLPD, and $\mathrm{Cov}_{95}$.}
\label{fig:fig3}
\end{figure*}
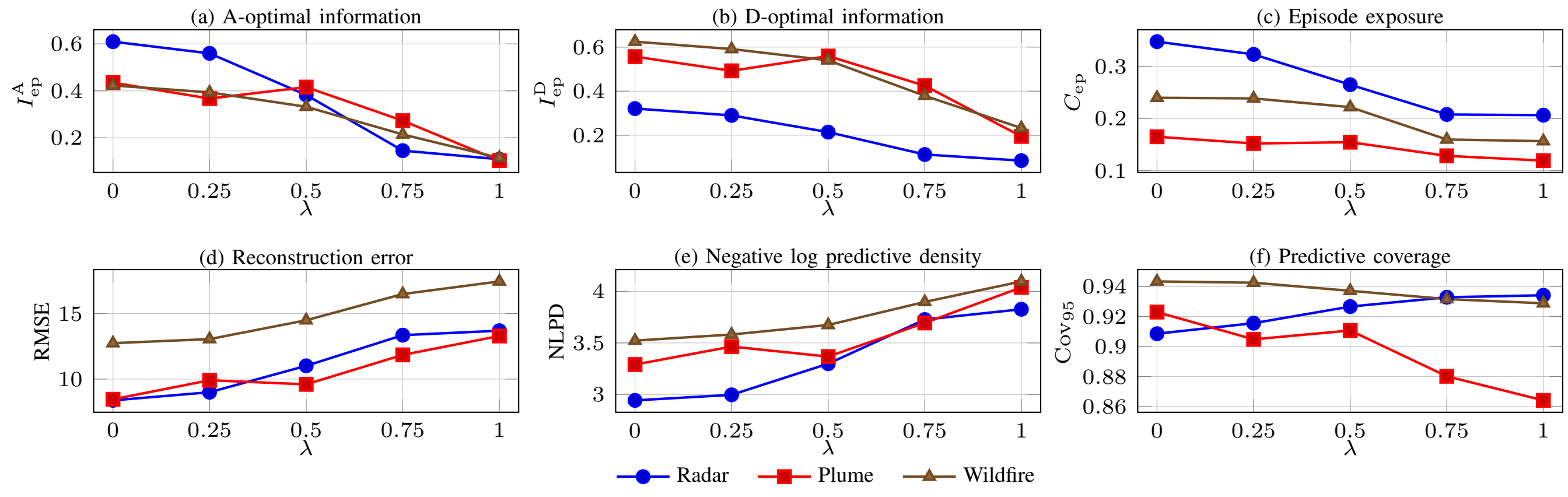
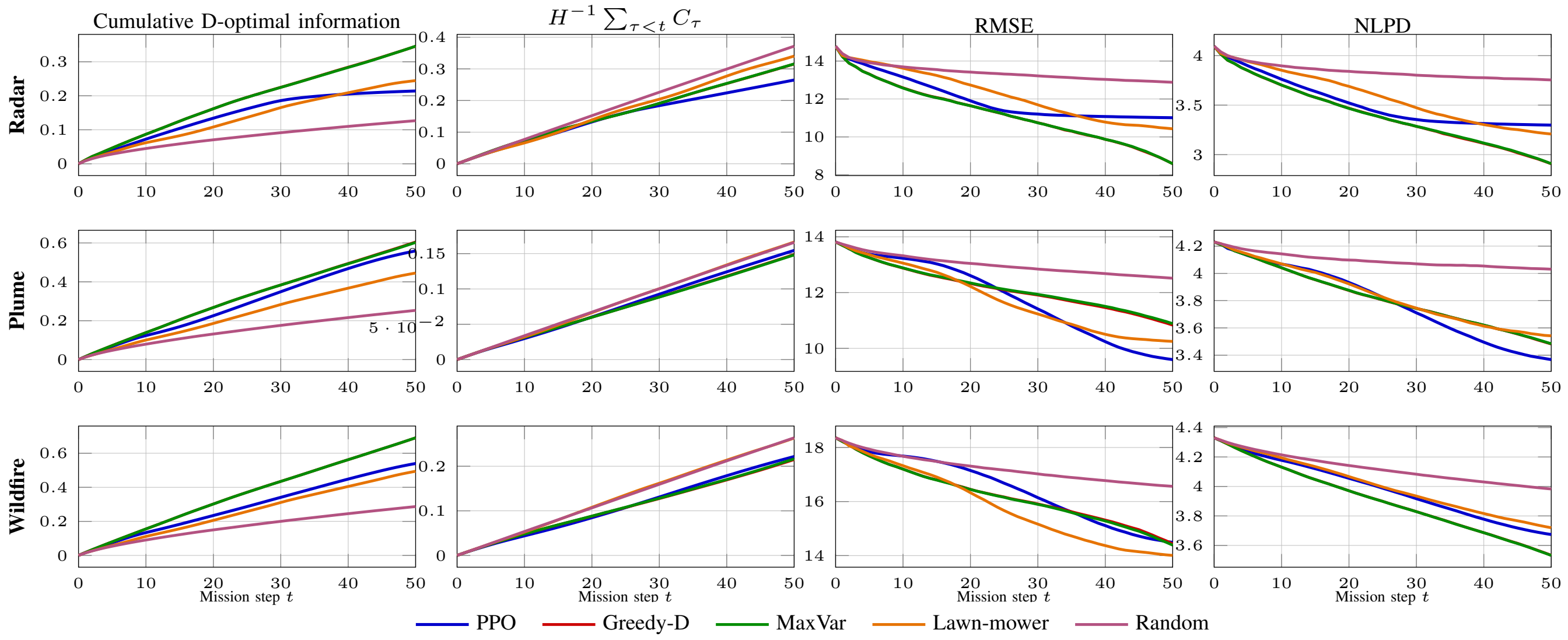
\begin{figure*}
\centering

\def\MissionLambda{0.5}

\def\RadarMissionCSV{data11/tables/mission/radar_d_lambda\MissionLambda.csv}
\def\ChemicalMissionCSV{data11/tables/mission/chemical_plume_d_lambda\MissionLambda.csv}
\def\WildfireMissionCSV{data11/tables/mission/wildfires_d_lambda\MissionLambda.csv}

\pgfplotsset{
    ppoStyle/.style={
        solid,
        no marks,
        line width=0.95pt,
        color=blue!80!black,
    },
    greedyStyle/.style={
        solid,
        no marks,
        line width=0.90pt,
        color=red!80!black,
    },
    maxvarStyle/.style={
        solid,
        no marks,
        line width=0.90pt,
        color=green!55!black,
    },
    lawnmowerStyle/.style={
        solid,
        no marks,
        line width=0.90pt,
        color=orange!90!black,
    },
    randomStyle/.style={
        solid,
        no marks,
        line width=0.90pt,
        color=magenta!70!black,
    },
}

\begin{tikzpicture}

\begin{groupplot}[
    group style={
        group size=4 by 3,
        horizontal sep=0.470cm,
        vertical sep=0.62cm,
    },
    width=0.215\textwidth,
    height=0.09\textwidth,
    scale only axis,
    xmin=0,
    xmax=50,
    ylabel style={xshift=3pt,yshift=-4pt},
    xlabel style={yshift=6pt},
    xtick={0,10,20,30,40,50},
    grid=major,
    grid style={line width=0.15pt},
    axis line style={line width=0.25pt},
    tick style={line width=0.25pt},
    tick label style={font=\tiny},
    label style={font=\tiny},
    title style={
        font=\scriptsize,
        yshift=-9pt,
    },
    legend style={
        font=\scriptsize,
        draw=none,
        legend columns=5,
        /tikz/every even column/.append style={column sep=6pt},
    },
    legend image post style={mark=none},
]


\nextgroupplot[
    title={Cumulative D-optimal information},
    legend to name=missionLegend,
]

\addplot+[ppoStyle]
table[x=step,y=ppoIMean,col sep=comma]{\RadarMissionCSV};
\addlegendentry{PPO}

\addplot+[greedyStyle]
table[x=step,y=greedyDIMean,col sep=comma]{\RadarMissionCSV};
\addlegendentry{Greedy-D}

\addplot+[maxvarStyle]
table[x=step,y=maxvarIMean,col sep=comma]{\RadarMissionCSV};
\addlegendentry{MaxVar}

\addplot+[lawnmowerStyle]
table[x=step,y=lawnmowerIMean,col sep=comma]{\RadarMissionCSV};
\addlegendentry{Lawn-mower}

\addplot+[randomStyle]
table[x=step,y=randomIMean,col sep=comma]{\RadarMissionCSV};
\addlegendentry{Random}

\nextgroupplot[
    title={$H^{-1}\sum_{\tau<t} C_\tau$},
]

\addplot+[ppoStyle]
table[x=step,y=ppoCMean,col sep=comma]{\RadarMissionCSV};

\addplot+[greedyStyle]
table[x=step,y=greedyDCMean,col sep=comma]{\RadarMissionCSV};

\addplot+[maxvarStyle]
table[x=step,y=maxvarCMean,col sep=comma]{\RadarMissionCSV};

\addplot+[lawnmowerStyle]
table[x=step,y=lawnmowerCMean,col sep=comma]{\RadarMissionCSV};

\addplot+[randomStyle]
table[x=step,y=randomCMean,col sep=comma]{\RadarMissionCSV};

\nextgroupplot[
    title={RMSE},
]

\addplot+[ppoStyle]
table[x=step,y=ppoRmseMean,col sep=comma]{\RadarMissionCSV};

\addplot+[greedyStyle]
table[x=step,y=greedyDRmseMean,col sep=comma]{\RadarMissionCSV};

\addplot+[maxvarStyle]
table[x=step,y=maxvarRmseMean,col sep=comma]{\RadarMissionCSV};

\addplot+[lawnmowerStyle]
table[x=step,y=lawnmowerRmseMean,col sep=comma]{\RadarMissionCSV};

\addplot+[randomStyle]
table[x=step,y=randomRmseMean,col sep=comma]{\RadarMissionCSV};

\nextgroupplot[
    title={NLPD},
]

\addplot+[ppoStyle]
table[x=step,y=ppoNlpdMean,col sep=comma]{\RadarMissionCSV};

\addplot+[greedyStyle]
table[x=step,y=greedyDNlpdMean,col sep=comma]{\RadarMissionCSV};

\addplot+[maxvarStyle]
table[x=step,y=maxvarNlpdMean,col sep=comma]{\RadarMissionCSV};

\addplot+[lawnmowerStyle]
table[x=step,y=lawnmowerNlpdMean,col sep=comma]{\RadarMissionCSV};

\addplot+[randomStyle]
table[x=step,y=randomNlpdMean,col sep=comma]{\RadarMissionCSV};


\nextgroupplot[]

\addplot+[ppoStyle]
table[x=step,y=ppoIMean,col sep=comma]{\ChemicalMissionCSV};

\addplot+[greedyStyle]
table[x=step,y=greedyDIMean,col sep=comma]{\ChemicalMissionCSV};

\addplot+[maxvarStyle]
table[x=step,y=maxvarIMean,col sep=comma]{\ChemicalMissionCSV};

\addplot+[lawnmowerStyle]
table[x=step,y=lawnmowerIMean,col sep=comma]{\ChemicalMissionCSV};

\addplot+[randomStyle]
table[x=step,y=randomIMean,col sep=comma]{\ChemicalMissionCSV};

\nextgroupplot[]

\addplot+[ppoStyle]
table[x=step,y=ppoCMean,col sep=comma]{\ChemicalMissionCSV};

\addplot+[greedyStyle]
table[x=step,y=greedyDCMean,col sep=comma]{\ChemicalMissionCSV};

\addplot+[maxvarStyle]
table[x=step,y=maxvarCMean,col sep=comma]{\ChemicalMissionCSV};

\addplot+[lawnmowerStyle]
table[x=step,y=lawnmowerCMean,col sep=comma]{\ChemicalMissionCSV};

\addplot+[randomStyle]
table[x=step,y=randomCMean,col sep=comma]{\ChemicalMissionCSV};

\nextgroupplot[]

\addplot+[ppoStyle]
table[x=step,y=ppoRmseMean,col sep=comma]{\ChemicalMissionCSV};

\addplot+[greedyStyle]
table[x=step,y=greedyDRmseMean,col sep=comma]{\ChemicalMissionCSV};

\addplot+[maxvarStyle]
table[x=step,y=maxvarRmseMean,col sep=comma]{\ChemicalMissionCSV};

\addplot+[lawnmowerStyle]
table[x=step,y=lawnmowerRmseMean,col sep=comma]{\ChemicalMissionCSV};

\addplot+[randomStyle]
table[x=step,y=randomRmseMean,col sep=comma]{\ChemicalMissionCSV};

\nextgroupplot[]

\addplot+[ppoStyle]
table[x=step,y=ppoNlpdMean,col sep=comma]{\ChemicalMissionCSV};

\addplot+[greedyStyle]
table[x=step,y=greedyDNlpdMean,col sep=comma]{\ChemicalMissionCSV};

\addplot+[maxvarStyle]
table[x=step,y=maxvarNlpdMean,col sep=comma]{\ChemicalMissionCSV};

\addplot+[lawnmowerStyle]
table[x=step,y=lawnmowerNlpdMean,col sep=comma]{\ChemicalMissionCSV};

\addplot+[randomStyle]
table[x=step,y=randomNlpdMean,col sep=comma]{\ChemicalMissionCSV};


\nextgroupplot[
    xlabel={Mission step $t$},
]

\addplot+[ppoStyle]
table[x=step,y=ppoIMean,col sep=comma]{\WildfireMissionCSV};

\addplot+[greedyStyle]
table[x=step,y=greedyDIMean,col sep=comma]{\WildfireMissionCSV};

\addplot+[maxvarStyle]
table[x=step,y=maxvarIMean,col sep=comma]{\WildfireMissionCSV};

\addplot+[lawnmowerStyle]
table[x=step,y=lawnmowerIMean,col sep=comma]{\WildfireMissionCSV};

\addplot+[randomStyle]
table[x=step,y=randomIMean,col sep=comma]{\WildfireMissionCSV};

\nextgroupplot[
    xlabel={Mission step $t$},
]

\addplot+[ppoStyle]
table[x=step,y=ppoCMean,col sep=comma]{\WildfireMissionCSV};

\addplot+[greedyStyle]
table[x=step,y=greedyDCMean,col sep=comma]{\WildfireMissionCSV};

\addplot+[maxvarStyle]
table[x=step,y=maxvarCMean,col sep=comma]{\WildfireMissionCSV};

\addplot+[lawnmowerStyle]
table[x=step,y=lawnmowerCMean,col sep=comma]{\WildfireMissionCSV};

\addplot+[randomStyle]
table[x=step,y=randomCMean,col sep=comma]{\WildfireMissionCSV};

\nextgroupplot[
    xlabel={Mission step $t$},
]

\addplot+[ppoStyle]
table[x=step,y=ppoRmseMean,col sep=comma]{\WildfireMissionCSV};

\addplot+[greedyStyle]
table[x=step,y=greedyDRmseMean,col sep=comma]{\WildfireMissionCSV};

\addplot+[maxvarStyle]
table[x=step,y=maxvarRmseMean,col sep=comma]{\WildfireMissionCSV};

\addplot+[lawnmowerStyle]
table[x=step,y=lawnmowerRmseMean,col sep=comma]{\WildfireMissionCSV};

\addplot+[randomStyle]
table[x=step,y=randomRmseMean,col sep=comma]{\WildfireMissionCSV};

\nextgroupplot[
    xlabel={Mission step $t$},
]

\addplot+[ppoStyle]
table[x=step,y=ppoNlpdMean,col sep=comma]{\WildfireMissionCSV};

\addplot+[greedyStyle]
table[x=step,y=greedyDNlpdMean,col sep=comma]{\WildfireMissionCSV};

\addplot+[maxvarStyle]
table[x=step,y=maxvarNlpdMean,col sep=comma]{\WildfireMissionCSV};

\addplot+[lawnmowerStyle]
table[x=step,y=lawnmowerNlpdMean,col sep=comma]{\WildfireMissionCSV};

\addplot+[randomStyle]
table[x=step,y=randomNlpdMean,col sep=comma]{\WildfireMissionCSV};

\end{groupplot}


\node[
    rotate=90,
    font=\scriptsize\bfseries
]
at ($(group c1r1.west)+(-0.70cm,0)$)
{Radar};

\node[
    rotate=90,
    font=\scriptsize\bfseries
]
at ($(group c1r2.west)+(-0.70cm,0)$)
{Plume};

\node[
    rotate=90,
    font=\scriptsize\bfseries
]
at ($(group c1r3.west)+(-0.70cm,0)$)
{Wildfire};


\node at
($(group c2r3.south)!0.5!(group c3r3.south)+(0,-0.65cm)$)
{\ref{missionLegend}};

\end{tikzpicture}

\vspace{-1.5em}
\caption{Mission-horizon comparison of PPO at $\lambda=\MissionLambda$ with non-learning baselines under D-optimal sensing. Columns show cumulative D-optimal information, normalized cumulative exposure, GP reconstruction RMSE, and NLPD for the radar, chemical-plume, and wildfire field families.}
\label{fig:fig4}

\end{figure*}

\begin{table}[t]
\centering
\small

\caption{PPO at $\lambda\in \{0,0.25,0.5,0.75,1\}$ compared with non-learning baselines under D-optimal sensing. Bold denotes the best value within each field.}

\label{tab:table2}

\pgfplotstableread[col sep=comma]
{data11/tables/final/radar_d_lambda0.csv}
\RadarZeroTable

\pgfplotstableread[col sep=comma]
{data11/tables/final/radar_d_lambda0.25.csv}
\RadarQuarterTable

\pgfplotstableread[col sep=comma]
{data11/tables/final/radar_d_lambda0.5.csv}
\RadarHalfTable

\pgfplotstableread[col sep=comma]
{data11/tables/final/radar_d_lambda0.75.csv}
\RadarThreeQuarterTable

\pgfplotstableread[col sep=comma]
{data11/tables/final/radar_d_lambda1.csv}
\RadarOneTable

\pgfplotstableread[col sep=comma]
{data11/tables/final/chemical_plume_d_lambda0.csv}
\ChemicalZeroTable

\pgfplotstableread[col sep=comma]
{data11/tables/final/chemical_plume_d_lambda0.25.csv}
\ChemicalQuarterTable

\pgfplotstableread[col sep=comma]
{data11/tables/final/chemical_plume_d_lambda0.5.csv}
\ChemicalHalfTable

\pgfplotstableread[col sep=comma]
{data11/tables/final/chemical_plume_d_lambda0.75.csv}
\ChemicalThreeQuarterTable

\pgfplotstableread[col sep=comma]
{data11/tables/final/chemical_plume_d_lambda1.csv}
\ChemicalOneTable

\pgfplotstableread[col sep=comma]
{data11/tables/final/wildfires_d_lambda0.csv}
\WildfireZeroTable

\pgfplotstableread[col sep=comma]
{data11/tables/final/wildfires_d_lambda0.25.csv}
\WildfireQuarterTable

\pgfplotstableread[col sep=comma]
{data11/tables/final/wildfires_d_lambda0.5.csv}
\WildfireHalfTable

\pgfplotstableread[col sep=comma]
{data11/tables/final/wildfires_d_lambda0.75.csv}
\WildfireThreeQuarterTable

\pgfplotstableread[col sep=comma]
{data11/tables/final/wildfires_d_lambda1.csv}
\WildfireOneTable

%
\newcommand{\ReadMethodMetric}[3]{%
  \def\FoundValue{nan}%
  \pgfplotstablegetrowsof{#3}%
  \pgfmathtruncatemacro{\LastRow}{\pgfplotsretval-1}%
  \pgfplotsforeachungrouped \r in {0,...,\LastRow}{%
    \pgfplotstablegetelem{\r}{method}\of#3%
    \edef\RowMethod{\pgfplotsretval}%
    \ifnum\pdfstrcmp{\RowMethod}{#1}=0
      \pgfplotstablegetelem{\r}{#2Mean}\of#3%
      \xdef\FoundValue{\pgfplotsretval}%
    \fi
  }%
}

%
\newcommand{\UpdateBestMethodMetric}[4]{%
  \ReadMethodMetric{#1}{#2}{#3}%
  \edef\ThisValue{\FoundValue}%
  \ifnum\pdfstrcmp{#4}{cov95}=0
    \pgfmathparse{abs(\ThisValue-0.95)}%
    \edef\ThisCompare{\pgfmathresult}%
    \pgfmathparse{min(\BestValue,\ThisCompare)}%
    \xdef\BestValue{\pgfmathresult}%
  \else
    \ifnum\pdfstrcmp{#4}{max}=0
      \pgfmathparse{max(\BestValue,\ThisValue)}%
      \xdef\BestValue{\pgfmathresult}%
    \else
      \pgfmathparse{min(\BestValue,\ThisValue)}%
      \xdef\BestValue{\pgfmathresult}%
    \fi
  \fi
}

%
%
\newcommand{\MethodMetricMean}[9]{%
  \def\CellMean{nan}%
  %
  \ReadMethodMetric{PPO}{#2}{#5}%
  \ifnum\pdfstrcmp{#4}{cov95}=0
    \pgfmathparse{abs(\FoundValue-0.95)}%
    \xdef\BestValue{\pgfmathresult}%
  \else
    \xdef\BestValue{\FoundValue}%
  \fi
  %
  \UpdateBestMethodMetric{PPO}{#2}{#5}{#4}%
  \UpdateBestMethodMetric{PPO}{#2}{#6}{#4}%
  \UpdateBestMethodMetric{PPO}{#2}{#7}{#4}%
  \UpdateBestMethodMetric{PPO}{#2}{#8}{#4}%
  \UpdateBestMethodMetric{PPO}{#2}{#9}{#4}%
  %
  \UpdateBestMethodMetric{Greedy-D}{#2}{#7}{#4}%
  \UpdateBestMethodMetric{MaxVar}{#2}{#7}{#4}%
  \UpdateBestMethodMetric{MinExposure}{#2}{#7}{#4}%
  \UpdateBestMethodMetric{LawnMower}{#2}{#7}{#4}%
  \UpdateBestMethodMetric{Random}{#2}{#7}{#4}%
  %
  \ReadMethodMetric{#1}{#2}{#3}%
  \xdef\CellMean{\FoundValue}%
  %
  \ifnum\pdfstrcmp{#4}{cov95}=0
    \pgfmathparse{abs(\CellMean-0.95)}%
    \xdef\CellCompare{\pgfmathresult}%
  \else
    \xdef\CellCompare{\CellMean}%
  \fi
  %
  \pgfmathparse{abs(\CellCompare-\BestValue)}%
  \ifdim\pgfmathresult pt<0.00001pt
    {\bfseries\boldmath\pgfmathprintnumber[fixed,precision=3]{\CellMean}}%
  \else
    \pgfmathprintnumber[fixed,precision=3]{\CellMean}%
  \fi
}

%
\newcommand{\BaselineRows}[6]{%
\multirow{11}{*}{#1}
& PPO ($\lambda=0$)
& \MethodMetricMean{PPO}{infoA}{#2}{max}{#2}{#3}{#4}{#5}{#6}
& \MethodMetricMean{PPO}{infoD}{#2}{max}{#2}{#3}{#4}{#5}{#6}
& \MethodMetricMean{PPO}{exposure}{#2}{min}{#2}{#3}{#4}{#5}{#6}
& \MethodMetricMean{PPO}{rmse}{#2}{min}{#2}{#3}{#4}{#5}{#6}
& \MethodMetricMean{PPO}{nlpd}{#2}{min}{#2}{#3}{#4}{#5}{#6}
\\

& PPO ($\lambda=0.25$)
& \MethodMetricMean{PPO}{infoA}{#3}{max}{#2}{#3}{#4}{#5}{#6}
& \MethodMetricMean{PPO}{infoD}{#3}{max}{#2}{#3}{#4}{#5}{#6}
& \MethodMetricMean{PPO}{exposure}{#3}{min}{#2}{#3}{#4}{#5}{#6}
& \MethodMetricMean{PPO}{rmse}{#3}{min}{#2}{#3}{#4}{#5}{#6}
& \MethodMetricMean{PPO}{nlpd}{#3}{min}{#2}{#3}{#4}{#5}{#6}
\\

& PPO ($\lambda=0.5$)
& \MethodMetricMean{PPO}{infoA}{#4}{max}{#2}{#3}{#4}{#5}{#6}
& \MethodMetricMean{PPO}{infoD}{#4}{max}{#2}{#3}{#4}{#5}{#6}
& \MethodMetricMean{PPO}{exposure}{#4}{min}{#2}{#3}{#4}{#5}{#6}
& \MethodMetricMean{PPO}{rmse}{#4}{min}{#2}{#3}{#4}{#5}{#6}
& \MethodMetricMean{PPO}{nlpd}{#4}{min}{#2}{#3}{#4}{#5}{#6}
\\

& PPO ($\lambda=0.75$)
& \MethodMetricMean{PPO}{infoA}{#5}{max}{#2}{#3}{#4}{#5}{#6}
& \MethodMetricMean{PPO}{infoD}{#5}{max}{#2}{#3}{#4}{#5}{#6}
& \MethodMetricMean{PPO}{exposure}{#5}{min}{#2}{#3}{#4}{#5}{#6}
& \MethodMetricMean{PPO}{rmse}{#5}{min}{#2}{#3}{#4}{#5}{#6}
& \MethodMetricMean{PPO}{nlpd}{#5}{min}{#2}{#3}{#4}{#5}{#6}
\\

& PPO ($\lambda=1$)
& \MethodMetricMean{PPO}{infoA}{#6}{max}{#2}{#3}{#4}{#5}{#6}
& \MethodMetricMean{PPO}{infoD}{#6}{max}{#2}{#3}{#4}{#5}{#6}
& \MethodMetricMean{PPO}{exposure}{#6}{min}{#2}{#3}{#4}{#5}{#6}
& \MethodMetricMean{PPO}{rmse}{#6}{min}{#2}{#3}{#4}{#5}{#6}
& \MethodMetricMean{PPO}{nlpd}{#6}{min}{#2}{#3}{#4}{#5}{#6}
\\

& Greedy-D\cite{chaloner1995bayesian}
& \MethodMetricMean{Greedy-D}{infoA}{#4}{max}{#2}{#3}{#4}{#5}{#6}
& \MethodMetricMean{Greedy-D}{infoD}{#4}{max}{#2}{#3}{#4}{#5}{#6}
& \MethodMetricMean{Greedy-D}{exposure}{#4}{min}{#2}{#3}{#4}{#5}{#6}
& \MethodMetricMean{Greedy-D}{rmse}{#4}{min}{#2}{#3}{#4}{#5}{#6}
& \MethodMetricMean{Greedy-D}{nlpd}{#4}{min}{#2}{#3}{#4}{#5}{#6}
\\

& MaxVar  \cite{krause2008near}
& \MethodMetricMean{MaxVar}{infoA}{#4}{max}{#2}{#3}{#4}{#5}{#6}
& \MethodMetricMean{MaxVar}{infoD}{#4}{max}{#2}{#3}{#4}{#5}{#6}
& \MethodMetricMean{MaxVar}{exposure}{#4}{min}{#2}{#3}{#4}{#5}{#6}
& \MethodMetricMean{MaxVar}{rmse}{#4}{min}{#2}{#3}{#4}{#5}{#6}
& \MethodMetricMean{MaxVar}{nlpd}{#4}{min}{#2}{#3}{#4}{#5}{#6}
\\

& MinExposure
& \MethodMetricMean{MinExposure}{infoA}{#4}{max}{#2}{#3}{#4}{#5}{#6}
& \MethodMetricMean{MinExposure}{infoD}{#4}{max}{#2}{#3}{#4}{#5}{#6}
& \MethodMetricMean{MinExposure}{exposure}{#4}{min}{#2}{#3}{#4}{#5}{#6}
& \MethodMetricMean{MinExposure}{rmse}{#4}{min}{#2}{#3}{#4}{#5}{#6}
& \MethodMetricMean{MinExposure}{nlpd}{#4}{min}{#2}{#3}{#4}{#5}{#6}
\\

& Lawn-Mower
& \MethodMetricMean{LawnMower}{infoA}{#4}{max}{#2}{#3}{#4}{#5}{#6}
& \MethodMetricMean{LawnMower}{infoD}{#4}{max}{#2}{#3}{#4}{#5}{#6}
& \MethodMetricMean{LawnMower}{exposure}{#4}{min}{#2}{#3}{#4}{#5}{#6}
& \MethodMetricMean{LawnMower}{rmse}{#4}{min}{#2}{#3}{#4}{#5}{#6}
& \MethodMetricMean{LawnMower}{nlpd}{#4}{min}{#2}{#3}{#4}{#5}{#6}
\\

& Random
& \MethodMetricMean{Random}{infoA}{#4}{max}{#2}{#3}{#4}{#5}{#6}
& \MethodMetricMean{Random}{infoD}{#4}{max}{#2}{#3}{#4}{#5}{#6}
& \MethodMetricMean{Random}{exposure}{#4}{min}{#2}{#3}{#4}{#5}{#6}
& \MethodMetricMean{Random}{rmse}{#4}{min}{#2}{#3}{#4}{#5}{#6}
& \MethodMetricMean{Random}{nlpd}{#4}{min}{#2}{#3}{#4}{#5}{#6}
\\
}

\renewcommand{\arraystretch}{0.35}
\setlength{\tabcolsep}{3.6pt}
\scriptsize

\begin{tabular}{llcccccc}
\toprule
Field
& Method
& $I_{\mathrm{ep}}^{\mathrm A}$
& $I_{\mathrm{ep}}^{\mathrm D}$
& $C_{\mathrm{ep}}$
& RMSE
& NLPD
\\
\midrule

\BaselineRows
{Radar}
{\RadarZeroTable}
{\RadarQuarterTable}
{\RadarHalfTable}
{\RadarThreeQuarterTable}
{\RadarOneTable}

\midrule

\BaselineRows
{Chem.}
{\ChemicalZeroTable}
{\ChemicalQuarterTable}
{\ChemicalHalfTable}
{\ChemicalThreeQuarterTable}
{\ChemicalOneTable}

\midrule

\BaselineRows
{Wildfire}
{\WildfireZeroTable}
{\WildfireQuarterTable}
{\WildfireHalfTable}
{\WildfireThreeQuarterTable}
{\WildfireOneTable}

\bottomrule
\end{tabular}%

\vspace{-2em}
\end{table}

\subsection{Comparison With Non-Learning Baselines}
\label{subsec:results_baselines}
Figure~\ref{fig:fig4} compares PPO at $\lambda=0.5$ with the non-learning baselines over the mission horizon, while Table~\ref{tab:table2} reports all PPO policies with different $\lambda$. Information-seeking PPO ($\lambda=0$) remains within $7.2\%$, $7.9\%$, and $9.6\%$ of the strongest D-optimal baseline for radar, plume, and wildfire, respectively, while achieving the lowest RMSE in all three fields. Conversely, high-$\lambda$ PPO achieves lower exposure than the information-driven baselines at the cost of higher RMSE.

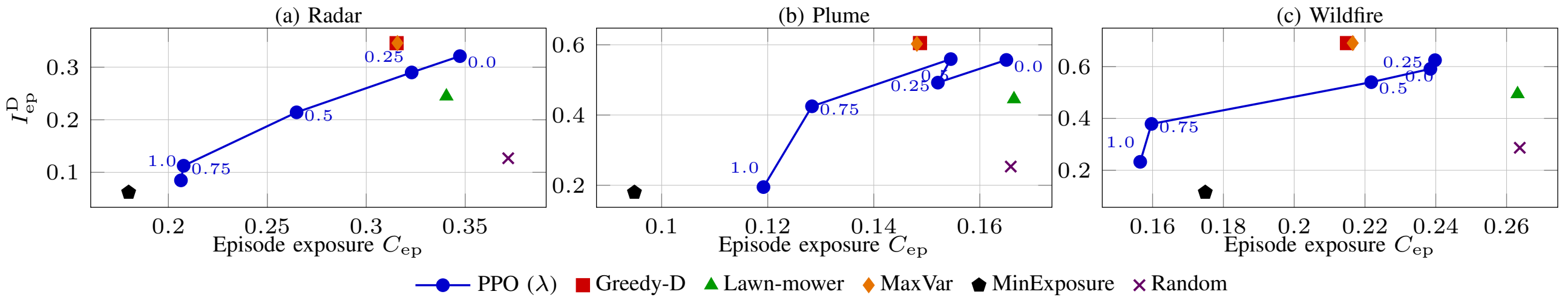
\begin{figure*}
\centering

\def\RadarTradeCSV{data11/tables/tradeoff/radar_d_tradeoff.csv}
\def\ChemicalTradeCSV{data11/tables/tradeoff/chemical_plume_d_tradeoff.csv}
\def\WildfireTradeCSV{data11/tables/tradeoff/wildfires_d_tradeoff.csv}

\def\RadarBaseCSV{data11/tables/tradeoff/radar_d_baselines.csv}
\def\ChemicalBaseCSV{data11/tables/tradeoff/chemical_plume_d_baselines.csv}
\def\WildfireBaseCSV{data11/tables/tradeoff/wildfires_d_baselines.csv}

\pgfplotsset{
    keep row 0/.style={
        x filter/.code={
            \ifnum\coordindex=0\else
                \def\pgfmathresult{}
            \fi
        },
    },
    keep row 1/.style={
        x filter/.code={
            \ifnum\coordindex=1\else
                \def\pgfmathresult{}
            \fi
        },
    },
    keep row 2/.style={
        x filter/.code={
            \ifnum\coordindex=2\else
                \def\pgfmathresult{}
            \fi
        },
    },
    keep row 3/.style={
        x filter/.code={
            \ifnum\coordindex=3\else
                \def\pgfmathresult{}
            \fi
        },
    },
    keep row 4/.style={
        x filter/.code={
            \ifnum\coordindex=4\else
                \def\pgfmathresult{}
            \fi
        },
    },
}

\begin{tikzpicture}

\begin{groupplot}[
    group style={
        group size=3 by 1,
        horizontal sep=0.55cm,
    },
    width=0.28\textwidth,
    height=0.11\textwidth,
    scale only axis,
    grid=major,
    ylabel style={xshift=3pt,yshift=-4pt},
    xlabel style={yshift=6pt},
    grid style={line width=0.15pt},
    axis line style={line width=0.25pt},
    tick style={line width=0.25pt},
    xlabel={Episode exposure $C_{\mathrm{ep}}$},
    tick label style={font=\scriptsize},
    label style={font=\scriptsize},
    title style={
        font=\scriptsize,
        yshift=-9pt,
    },
    legend style={
        font=\scriptsize,
        draw=none,
        legend columns=6,
        /tikz/every even column/.append style={column sep=4pt},
    },
]

\nextgroupplot[
    title={(a) Radar},
    ylabel={$I_{\mathrm{ep}}^{\mathrm D}$},
    legend to name=paretoLegend,
]

\addplot[
    solid,
    color=blue!80!black,
    line width=0.65pt,
    mark=*,
    mark size=1.8pt,
    nodes near coords={$\pgfplotspointmeta$},
    point meta=explicit symbolic,
    every node near coord/.append style={
        font=\tiny,
        inner sep=1pt,
        /utils/exec={
            \ifcase\coordindex\relax
                \pgfkeysalso{anchor=north west,xshift=1pt,yshift=1pt}%
            \or
                \pgfkeysalso{anchor=south east,xshift=-1pt,yshift=2pt}%
            \or
                \pgfkeysalso{anchor=north west,xshift=1pt,yshift=2pt}%
            \or
                \pgfkeysalso{anchor=north west,xshift=1pt,yshift=2pt}%
            \or
                \pgfkeysalso{anchor=north east,xshift=0pt,yshift=9pt}%
            \fi
        },
    },
]
table[
    col sep=comma,
    x=ppoCMean,
    y=ppoIMean,
    meta=lambda
]{\RadarTradeCSV};
\addlegendentry{PPO ($\lambda$)}

\addplot[
    only marks,
    color=red!80!black,
    mark=square*,
    mark size=2pt,
    keep row 0,
]
table[
    col sep=comma,
    x=CMean,
    y=IMean
]{\RadarBaseCSV};
\addlegendentry{Greedy-D}

\addplot[
    only marks,
    color=green!60!black,
    mark=triangle*,
    mark size=2.2pt,
    keep row 1,
]
table[
    col sep=comma,
    x=CMean,
    y=IMean
]{\RadarBaseCSV};
\addlegendentry{Lawn-mower}

\addplot[
    only marks,
    color=orange!90!black,
    mark=diamond*,
    mark size=2.2pt,
    keep row 2,
]
table[
    col sep=comma,
    x=CMean,
    y=IMean
]{\RadarBaseCSV};
\addlegendentry{MaxVar}

\addplot[
    only marks,
    color=black,
    mark=pentagon*,
    mark size=2.2pt,
    keep row 3,
]
table[
    col sep=comma,
    x=CMean,
    y=IMean
]{\RadarBaseCSV};
\addlegendentry{MinExposure}

\addplot[
    only marks,
    color=violet!80!black,
    mark=x,
    mark size=2.4pt,
    line width=0.7pt,
    keep row 4,
]
table[
    col sep=comma,
    x=CMean,
    y=IMean
]{\RadarBaseCSV};
\addlegendentry{Random}

\nextgroupplot[
    title={(b) Plume},
]

\addplot[
    solid,
    color=blue!80!black,
    line width=0.65pt,
    mark=*,
    mark size=1.8pt,
    nodes near coords={$\pgfplotspointmeta$},
    point meta=explicit symbolic,
    every node near coord/.append style={
        font=\tiny,
        inner sep=1pt,
        /utils/exec={
            \ifcase\coordindex\relax
                \pgfkeysalso{anchor=north west,xshift=1pt,yshift=1pt}%
            \or
                \pgfkeysalso{anchor=north east,xshift=-1pt,yshift=2pt}%
            \or
                \pgfkeysalso{anchor=south east,xshift=1pt,yshift=-8pt}%
            \or
                \pgfkeysalso{anchor=north west,xshift=1pt,yshift=2pt}%
            \or
                \pgfkeysalso{anchor=north east,xshift=0pt,yshift=9pt}%
            \fi
        },
    },
]
table[
    col sep=comma,
    x=ppoCMean,
    y=ppoIMean,
    meta=lambda
]{\ChemicalTradeCSV};

\addplot[
    only marks,
    color=red!80!black,
    mark=square*,
    mark size=2pt,
    keep row 0,
]
table[
    col sep=comma,
    x=CMean,
    y=IMean
]{\ChemicalBaseCSV};

\addplot[
    only marks,
    color=green!60!black,
    mark=triangle*,
    mark size=2.2pt,
    keep row 1,
]
table[
    col sep=comma,
    x=CMean,
    y=IMean
]{\ChemicalBaseCSV};

\addplot[
    only marks,
    color=orange!90!black,
    mark=diamond*,
    mark size=2.2pt,
    keep row 2,
]
table[
    col sep=comma,
    x=CMean,
    y=IMean
]{\ChemicalBaseCSV};

\addplot[
    only marks,
    color=black,
    mark=pentagon*,
    mark size=2.2pt,
    keep row 3,
]
table[
    col sep=comma,
    x=CMean,
    y=IMean
]{\ChemicalBaseCSV};

\addplot[
    only marks,
    color=violet!80!black,
    mark=x,
    mark size=2.4pt,
    line width=0.7pt,
    keep row 4,
]
table[
    col sep=comma,
    x=CMean,
    y=IMean
]{\ChemicalBaseCSV};

\nextgroupplot[
    title={(c) Wildfire},
]

\addplot[
    solid,
    color=blue!80!black,
    line width=0.65pt,
    mark=*,
    mark size=1.8pt,
    nodes near coords={$\pgfplotspointmeta$},
    point meta=explicit symbolic,
    every node near coord/.append style={
        font=\tiny,
        inner sep=1pt,
        /utils/exec={
            \ifcase\coordindex\relax
                \pgfkeysalso{anchor=north east,xshift=1pt,yshift=-2pt}%
            \or
                \pgfkeysalso{anchor=north east,xshift=-1pt,yshift=5pt}%
            \or
                \pgfkeysalso{anchor=north west,xshift=1pt,yshift=1pt}%
            \or
                \pgfkeysalso{anchor=north west,xshift=1pt,yshift=2pt}%
            \or
                \pgfkeysalso{anchor=north west,xshift=-12pt,yshift=9pt}%
            \fi
        },
    },
]
table[
    col sep=comma,
    x=ppoCMean,
    y=ppoIMean,
    meta=lambda
]{\WildfireTradeCSV};

\addplot[
    only marks,
    color=red!80!black,
    mark=square*,
    mark size=2pt,
    keep row 0,
]
table[
    col sep=comma,
    x=CMean,
    y=IMean
]{\WildfireBaseCSV};

\addplot[
    only marks,
    color=green!60!black,
    mark=triangle*,
    mark size=2.2pt,
    keep row 1,
]
table[
    col sep=comma,
    x=CMean,
    y=IMean
]{\WildfireBaseCSV};

\addplot[
    only marks,
    color=orange!90!black,
    mark=diamond*,
    mark size=2.2pt,
    keep row 2,
]
table[
    col sep=comma,
    x=CMean,
    y=IMean
]{\WildfireBaseCSV};

\addplot[
    only marks,
    color=black,
    mark=pentagon*,
    mark size=2.2pt,
    keep row 3,
]
table[
    col sep=comma,
    x=CMean,
    y=IMean
]{\WildfireBaseCSV};

\addplot[
    only marks,
    color=violet!80!black,
    mark=x,
    mark size=2.4pt,
    line width=0.7pt,
    keep row 4,
]
table[
    col sep=comma,
    x=CMean,
    y=IMean
]{\WildfireBaseCSV};

\end{groupplot}

\node at ($(group c2r1.south)+(0,-0.85cm)$)
{\ref{paretoLegend}};

\end{tikzpicture}

\vspace{-1.5em}
\caption{D-optimal information--exposure trade-off for PPO and the non-learning baselines across the three field families. PPO markers indicate $\lambda\in\{0,0.25,0.5,0.75,1\}$.}
\label{fig:fig5}

\end{figure*}
\begin{figure*}
\centering


\def\RadarTrainZero{data11/training/radar_d_lambda0_N3.csv}
\def\RadarTrainQuarter{data11/training/radar_d_lambda0.25_N3.csv}
\def\RadarTrainHalf{data11/training/radar_d_lambda0.5_N3.csv}
\def\RadarTrainThreeQuarter{data11/training/radar_d_lambda0.75_N3.csv}
\def\RadarTrainOne{data11/training/radar_d_lambda1_N3.csv}

\def\ChemicalTrainZero{data11/training/chemical_plume_d_lambda0_N3.csv}
\def\ChemicalTrainQuarter{data11/training/chemical_plume_d_lambda0.25_N3.csv}
\def\ChemicalTrainHalf{data11/training/chemical_plume_d_lambda0.5_N3.csv}
\def\ChemicalTrainThreeQuarter{data11/training/chemical_plume_d_lambda0.75_N3.csv}
\def\ChemicalTrainOne{data11/training/chemical_plume_d_lambda1_N3.csv}

\def\WildfireTrainZero{data11/training/wildfires_d_lambda0_N3.csv}
\def\WildfireTrainQuarter{data11/training/wildfires_d_lambda0.25_N3.csv}
\def\WildfireTrainHalf{data11/training/wildfires_d_lambda0.5_N3.csv}
\def\WildfireTrainThreeQuarter{data11/training/wildfires_d_lambda0.75_N3.csv}
\def\WildfireTrainOne{data11/training/wildfires_d_lambda1_N3.csv}

\newcommand{\TrainCurve}[2]{%
\addplot+[
    solid,
    no marks,
    line width=0.75pt,
    color=#2
]
table[
    col sep=comma,
    x=episode,
    y=mean
]{#1};
}

\begin{tikzpicture}

\begin{groupplot}[
    group style={
        group size=3 by 1,
        horizontal sep=0.55cm,
    },
    width=0.28\textwidth,
    height=0.11\textwidth,
    scale only axis,
        ylabel style={xshift=3pt,yshift=-4pt},
    xlabel style={yshift=6pt},
    grid=major,
    grid style={line width=0.15pt},
    axis line style={line width=0.25pt},
    tick style={line width=0.25pt},
    xlabel={Training episode},
    tick label style={font=\scriptsize},
    label style={font=\scriptsize},
    title style={
        font=\scriptsize,
        yshift=-9pt,
    },
    legend style={
        font=\scriptsize,
        draw=none,
        legend columns=5,
        /tikz/every even column/.append style={column sep=6pt},
    },
]


\nextgroupplot[
    title={(a) Radar},
    ylabel={Evaluation return},
    legend to name=trainingLegend,
]

\TrainCurve{\RadarTrainZero}{blue!80!black}
\addlegendentry{$\lambda=0$}

\TrainCurve{\RadarTrainQuarter}{red!80!black}
\addlegendentry{$\lambda=0.25$}

\TrainCurve{\RadarTrainHalf}{green!55!black}
\addlegendentry{$\lambda=0.5$}

\TrainCurve{\RadarTrainThreeQuarter}{orange!90!black}
\addlegendentry{$\lambda=0.75$}

\TrainCurve{\RadarTrainOne}{magenta!70!black}
\addlegendentry{$\lambda=1$}


\nextgroupplot[
    title={(b) Plume},
]

\TrainCurve{\ChemicalTrainZero}{blue!80!black}
\TrainCurve{\ChemicalTrainQuarter}{red!80!black}
\TrainCurve{\ChemicalTrainHalf}{green!55!black}
\TrainCurve{\ChemicalTrainThreeQuarter}{orange!90!black}
\TrainCurve{\ChemicalTrainOne}{magenta!70!black}


\nextgroupplot[
    title={(c) Wildfire},
]

\TrainCurve{\WildfireTrainZero}{blue!80!black}
\TrainCurve{\WildfireTrainQuarter}{red!80!black}
\TrainCurve{\WildfireTrainHalf}{green!55!black}
\TrainCurve{\WildfireTrainThreeQuarter}{orange!90!black}
\TrainCurve{\WildfireTrainOne}{magenta!70!black}

\end{groupplot}


\node at ($(group c2r1.south)+(0,-0.9cm)$)
{\ref{trainingLegend}};

\end{tikzpicture}
\vspace{-1.5em}
\caption{PPO evaluation return during training under D-optimal sensing for $\lambda\in\{0,0.25,0.5,0.75,1\}$. Curves assess convergence within each configuration; return magnitudes are not compared across $\lambda$ because the scalarized reward changes with $\lambda$.}
\label{fig:fig6}

\end{figure*}
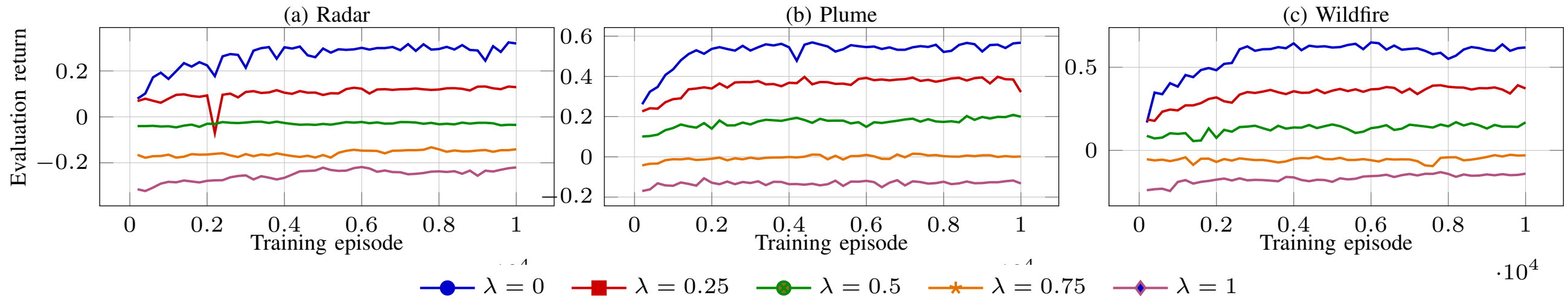

\subsection{Information-Exposure Trade-off}
\label{subsec:results_pareto}
The near-linear cumulative curves in Figure~\ref{fig:fig4} indicate per-step information and exposure rates, and their differing slopes show how each planner allocates the finite sensing budget.

\subsection{Training Convergence}
\label{subsec:results_training_gp}
Figure~\ref{fig:fig6} shows convergence toward distinct policy regimes for each $\lambda$; absolute returns are not comparable because the scalarized reward changes with $\lambda$. Together with Figures~\ref{fig:fig3} and~\ref{fig:fig5}, these results indicate distinct information-exposure operating regimes as $\lambda$ varies.

\section{Discussion}
\label{sec:discussion}
The results reveal two complementary operating regimes. When information dominates, PPO remains within $10\%$ of the strongest D-optimal baseline in every field while achieving the lowest RMSE, showing that greater covariance reduction does not necessarily yield lower reconstruction error. As $\lambda$ increases, PPO generally trades information for reduced exposure. \emph{MinExposure} attains lower exposure in radar and plume, whereas PPO attains lower exposure in wildfire. PPO nevertheless retains greater information than \emph{MinExposure}. Thus, \emph{Greedy-D} and \emph{MaxVar} remain effective for covariance reduction, while PPO provides a tunable information-exposure trade-off.

\section{Conclusion}
\label{sec:conclusion}
This paper formulated exposure-aware IPP as a finite-horizon GP belief-dependent decision-making problem and demonstrated the ability to learn policies using PPO that balance the exposure and information objectives. Across three field families, information-seeking PPO remained within $10\%$ of the strongest D-optimal baseline while achieving the lowest RMSE, whereas exposure-weighted policies reduced mean exposure by up to $40.6\%$ relative to information-seeking PPO. Thus, EA-IPP provides a controllable operating envelope between informative sensing and cumulative exposure. Future work will focus on extending the framework to spatio-temporal scalar fields, incorporating adaptive GP models, and deployment across heterogeneous robotic fleets.

\bibliographystyle{IEEETrans.bst}
\small
\bibliography{muzaffarICRA2,scc,sccmaster}

@InProceedings{SCC.Singh.Krause.ea2007,
  author    = {Singh, Amarjeet and Krause, Andreas and Guestrin, Carlos and Kaiser, William and Batalin, Maxim},
  booktitle = {Proc. Int. Joint Conf. Artif. Intell.},
  title     = {Efficient planning of informative paths for multiple robots},
  year      = {2007},
  pages     = {2204–2211},
  numpages  = {8},
}

@Article{SCC.Popovic.VidalCalleja.ea2020,
  author  = {Popovi\'c, Marija and Vidal-Calleja, Teresa and Hitz, Gregory and Chung, Jen Jen and Sa, Inkyu and Siegwart, Roland and Nieto, Juan},
  journal = {Auton. Robot.},
  title   = {An Informative Path Planning Framework for {UAV}-Based Terrain Monitoring},
  year    = {2020},
  number  = {6},
  pages   = {889--911},
  volume  = {44},
  doi     = {10.1007/s10514-020-09903-2},
}

@Article{SCC.Hollinger.Sukhatme2014,
  author  = {Geoffrey A. Hollinger and Gaurav S. Sukhatme},
  journal = {Int. J. Robot. Res.},
  title   = {Sampling-based robotic information gathering algorithms},
  year    = {2014},
  number  = {9},
  pages   = {1271-1287},
  volume  = {33},
  doi     = {10.1177/0278364914533443},
}

@inproceedings{SCC.Arora.Scherer2017,
  author    = {Arora, Sankalp and Scherer, Sebastian},
  title     = {Randomized Algorithm for Informative Path Planning with Budget Constraints},
  booktitle = {Proc. IEEE Int. Conf. Robot. Autom. },
  pages     = {4997--5004},
  year      = {2017},
  doi       = {10.1109/ICRA.2017.7989582}
}

@inproceedings{SCC.Binney.Sukhatme2012,
  author    = {Binney, Jonathan and Sukhatme, Gaurav S.},
  title     = {Branch and Bound for Informative Path Planning},
  booktitle = {Proc. IEEE Int. Conf. Robot. Autom.},
  pages     = {2147--2154},
  year      = {2012},
  doi       = {10.1109/ICRA.2012.6224902}
}

@article{hitz2017adaptive,
  author  = {Hitz, Gregory and Galceran, Enric and Garneau, Marie-\`Eve and Pomerleau, Fran\c{c}ois and Siegwart, Roland},
  title   = {Adaptive Continuous-Space Informative Path Planning for Online Environmental Monitoring},
  journal = {J. Field Robot.},
  volume  = {34},
  number  = {8},
  pages   = {1427--1449},
  year    = {2017},
  doi     = {10.1002/rob.21722}
}

@inproceedings{marchant2014bayesian,
  author    = {Marchant, Roman and Ramos, Fabio},
  title     = {Bayesian Optimisation for Informative Continuous Path Planning},
  booktitle = {Proc. IEEE Int. Conf. Robot. Autom.  },
  pages     = {6136--6143},
  year      = {2014},
  doi       = {10.1109/ICRA.2014.6907763}
}

@book{rasmussen2006gaussian,
  author    = {Rasmussen, Carl Edward and Williams, Christopher K. I.},
  title     = {Gaussian Processes for Machine Learning},
  publisher = {MIT Press},
  address   = {Cambridge, MA},
  year      = {2006},
  isbn      = {026218253X}
}

@article{krause2008near,
  author  = {Krause, Andreas and Singh, Ajit and Guestrin, Carlos},
  title   = {Near-Optimal Sensor Placements in {Gaussian} Processes: Theory, Efficient Algorithms and Empirical Studies},
  journal = {J. Mach. Learn. Res.},
  volume  = {9},
  pages   = {235--284},
  year    = {2008}
}

@article{chaloner1995bayesian,
  author  = {Chaloner, Kathryn and Verdinelli, Isabella},
  title   = {Bayesian Experimental Design: A Review},
  journal = {Stat. Sci.},
  volume  = {10},
  number  = {3},
  pages   = {273--304},
  year    = {1995},
  doi     = {10.1214/ss/1177009939}
}

@article{popovic2024survey,
  author  = {Popovi\'c, Marija and Ott, Joshua and R\"uckin, Julius and Kochenderfer, Mykel J.},
  title   = {Learning-Based Methods for Adaptive Informative Path Planning},
  journal = {Robot. Auton. Syst.},
  volume  = {179},
  pages   = {104727},
  year    = {2024},
  doi     = {10.1016/j.robot.2024.104727}
}

@article{yanes2021multiagent,
  author  = {Yanes Luis, Samuel and Guti\'errez Reina, Daniel and Toral Mar\'in, Sergio L.},
  title   = {A Multiagent Deep Reinforcement Learning Approach for Path Planning in Autonomous Surface Vehicles: The {Ypacara\'i} Lake Patrolling Case},
  journal = {IEEE Access},
  volume  = {9},
  pages   = {17084--17099},
  year    = {2021},
  doi     = {10.1109/ACCESS.2021.3053348}
}

@Article{SCC.YanesLuis.Shutin.ea2024,
  author  = {Yanes Luis, Samuel and Shutin, Dmitriy and Marchal Gómez, Juan and Gutiérrez Reina, Daniel and Toral Marín, Sergio},
  journal = {Adv. Intell. Syst.},
  title   = {Deep Reinforcement Multiagent Learning Framework for Information Gathering with Local {G}aussian Processes for Water Monitoring},
  year    = {2024},
  number  = {8},
  pages   = {2300850},
  volume  = {6},
  doi     ={10.1002/aisy.202300850},
}

@article{garcia2015survey,
  author  = {Garc\'ia, Javier and Fern\'andez, Fernando},
  title   = {A Comprehensive Survey on Safe Reinforcement Learning},
  journal = {J. Mach. Learn. Res.},
  volume  = {16},
  pages   = {1437--1480},
  year    = {2015}
}

@inproceedings{achiam2017cpo,
  author    = {Achiam, Joshua and Held, David and Tamar, Aviv and Abbeel, Pieter},
  title     = {Constrained Policy Optimization},
  booktitle = {Proc. Int. Conf. Mach. Learn.},
  volume    = {70},
  pages     = {22--31},
  year      = {2017},
  publisher = {PMLR}
}

@inproceedings{sui2015safe,
  author    = {Sui, Yanan and Gotovos, Alkis and Burdick, Joel W. and Krause, Andreas},
  title     = {Safe Exploration for Optimization with {Gaussian} Processes},
  booktitle = {Proc. Int. Conf. Mach. Learn.},
  volume    = {37},
  pages     = {997--1005},
  year      = {2015},
  publisher = {PMLR}
}

@inproceedings{turchetta2016safe,
  author    = {Turchetta, Matteo and Berkenkamp, Felix and Krause, Andreas},
  title     = {Safe Exploration in Finite {Markov} Decision Processes with {Gaussian} Processes},
  booktitle = {Proc. Adv. Neural Inf. Process. Syst.},
  volume    = {29},
  pages     = {4312--4320},
  year      = {2016}
}

@inproceedings{Qureshi2024ScalarFieldMapping,
  author    = {Qureshi, Muzaffar and Ogri, Tochukwu Elijah and Bell, Zachary I. and Kamalapurkar, Rushikesh},
  title     = {Scalar Field Mapping with Adaptive High-Intensity Region Avoidance},
  booktitle = {Proc. IEEE Conf. Control Technol. Appl.},
  address   = {Newcastle upon Tyne, UK},
  pages     = {388--393},
  month     = aug,
  year      = {2024},
  doi       = {10.1109/CCTA60707.2024.10666509}
}

@misc{Qureshi2026HoughTransform,
  author        = {Qureshi, Muzaffar and Satharasi, Trivikram and Ogri, Tochukwu E. and Volle, Kyle and Kamalapurkar, Rushikesh},
  title         = {A {Hough} Transform Approach to Safety-Aware Scalar Field Mapping Using {Gaussian} Processes},
  year          = {2026},
    howpublished = {{arXiv}:2604.20799},
}

@misc{schulman2017ppo,
  author        = {Schulman, John and Wolski, Filip and Dhariwal, Prafulla and Radford, Alec and Klimov, Oleg},
  title         = {Proximal Policy Optimization Algorithms},
  year          = {2017},
  howpublished = {{arXiv}:1707.06347},
}

@inproceedings{schulman2016gae,
  author    = {Schulman, John and Moritz, Philipp and Levine, Sergey and Jordan, Michael and Abbeel, Pieter},
  title     = {High-Dimensional Continuous Control Using Generalized Advantage Estimation},
  booktitle = {Int. Conf. Learn. Represent.},
  year      = {2016}
}

@Article{muzaffar.MacKay1992,
  author  = {MacKay, David J. C.},
  journal = {Neural Comput.},
  title   = {Information-Based Objective Functions for Active Data Selection},
  year    = {1992},
  month   = jul,
  number  = {4},
  pages   = {590--604},
  volume  = {4},
  doi     = {10.1162/neco.1992.4.4.590},
}

\end{document}